\documentclass[10pt,journal,compsoc]{IEEEtran}

\usepackage[nocompress]{cite}
\usepackage{url}
\usepackage{ragged2e}
\usepackage{epsfig}
\usepackage{graphicx}
\usepackage{amsmath,amssymb} % define this before the line numbering.
\usepackage{mathptmx}      % use Times fonts if available on your TeX system
\usepackage{subfigure}
\usepackage{algorithm}
\usepackage{algorithmicx}
\usepackage{algpseudocode}
\usepackage{graphics}
\usepackage{mathrsfs}
\usepackage{threeparttable}
\usepackage{color}
\usepackage[normalem]{ulem}
\usepackage{multirow}
\usepackage{float}
\usepackage{amsfonts}
\usepackage{bm}
\usepackage{array}
\usepackage[table]{xcolor}
\usepackage{colortbl}
\usepackage{pifont}
\usepackage{diagbox}
\usepackage{rotating}
\usepackage{booktabs}
\usepackage{overpic}
\usepackage{textcomp}
\usepackage{contour}
\usepackage{enumitem}
\usepackage{stfloats}
\usepackage{csquotes}
\usepackage{wasysym}
\usepackage[american]{babel}
\usepackage{microtype}

\definecolor{citeGreen}{RGB}{69, 137, 51}
\usepackage[colorlinks=true,linkcolor=black,citecolor=citeGreen]{hyperref}

\usepackage{silence}
\def\ie{\emph{i.e.}}
\def\eg{\emph{e.g.}}
\def\etc{\emph{etc}}
\def\etal{{\em et al.~}}

\newcommand{\sArt}{state-of-the-art }

\definecolor{bblue}{rgb}{0,150,230}
\definecolor{mygray}{gray}{.92}

\newcommand{\figref}[1]{Fig.~\ref{#1}}
\newcommand{\tabref}[1]{Table~\ref{#1}}
\newcommand{\eqnref}[1]{(Eq.~\ref{#1})}
\newcommand{\secref}[1]{$\S$ \ref{#1}}
\newcommand{\AddText}[3]{\put(#1,#2){\contour{white}{\textbf{\textcolor{black}{#3}}}}}
\newcommand{\AddAttr}[3]{\put(#1,#2){\contour{black}{\textbf{\textcolor{white}{#3}}}}}

\newcommand{\trb}[1]{\textbf{\textcolor{red}{#1}}}
\newcommand{\tgb}[1]{\textbf{\textcolor{green}{#1}}}
\newcommand{\tbb}[1]{\textbf{\textcolor{blue}{#1}}}
\newcommand{\supp}[1]{#1}

\def\ourdataset{\textit{SOC}}

\graphicspath{{./Imgs/}}
\DeclareGraphicsExtensions{.jpg,.pdf,.png}

\RequirePackage{silence}
\begin{document}

\title{Salient Objects in Clutter}

\author{Deng-Ping~Fan,~
        Jing Zhang,~
        Gang~Xu,~
        Ming-Ming Cheng,
        and Ling Shao,~\IEEEmembership{Fellow,~IEEE}\\
% <-this % stops a space
\IEEEcompsocitemizethanks{
\IEEEcompsocthanksitem Deng-Ping Fan, Gang Xu and Ming-Ming Cheng are with the CS, Nankai University,
Tianjin, China. %(E-mail: dengpfan@gmail.com; gangxu@mail.nankai.edu.cn; cmm@nankai.edu.cn)
\IEEEcompsocthanksitem Jing Zhang is with Research School of Engineering, the Australian National University, ACRV, DATA61-CSIRO. %(Email: zjnwpu@gmail.com)
\IEEEcompsocthanksitem Ling Shao is with Terminus Group, Beijing, China. %(E-mail: ling.shao@ieee.org)
%\IEEEcompsocthanksitem A preliminary version of this work appeared in ECCV 2018~\cite{fan2018salient}.
\IEEEcompsocthanksitem The major part of this work was done
in Nankai University.
\IEEEcompsocthanksitem Ming-Ming Cheng is the corresponding author (cmm@nankai.edu.cn).
\IEEEcompsocthanksitem This work is funded by the National Key Research and Development Program
of China (No. 2018AAA0100400)
and NSFC (NO. 61922046 and NO. 61929104).
}% <-this % stops an unwanted space
%\thanks{Manuscript received XXX; revised XXX.}
}

\markboth{IEEE TRANSACTIONS ON PATTERN ANALYSIS AND MACHINE INTELLIGENCE}%
{Fan \MakeLowercase{\textit{et al.}}: Salient Objects in Clutter}

\IEEEtitleabstractindextext{%
\begin{abstract}
\justifying
   In this paper, we identify and address a serious design bias of existing salient object detection (SOD) datasets, which unrealistically assume that each image should
   contain at least one clear and uncluttered salient object.
   This \textit{design bias} has led to a saturation in performance for \sArt~SOD models when
   evaluated on existing datasets. However, these models are still far from satisfactory when applied to real-world scenes.
   Based on our analyses,
   we propose a new high-quality dataset
   and update the previous saliency benchmark. Specifically, our dataset, called Salient Objects in Clutter~\textbf{(SOC)}, includes images with both salient and non-salient objects from several common
   object categories.
   In addition to object category annotations, each salient image is
   accompanied by attributes
   that reflect common challenges in common scenes, which can help provide deeper insight into the SOD problem.
   Further, with a given saliency encoder, \eg, the backbone network, existing saliency models are designed to achieve mapping from the training image set to the training ground-truth set. We therefore argue that improving the dataset can yield higher performance gains than focusing only on the decoder design.
   With this in mind, we investigate several dataset-enhancement strategies, including
   label smoothing to implicitly emphasize salient boundaries, random image augmentation to adapt saliency models to various scenarios, and self-supervised learning as a regularization strategy to learn from small datasets. Our extensive results demonstrate the effectiveness of these tricks.
   We also provide a comprehensive benchmark for SOD, which can be
   found in our repository: \url{https://github.com/DengPingFan/SODBenchmark}.
\end{abstract}

\begin{IEEEkeywords}
Salient object detection, SOD, SOC, survey, dataset, benchmark.
\end{IEEEkeywords}}

\maketitle

\IEEEdisplaynontitleabstractindextext

\IEEEpeerreviewmaketitle

\IEEEraisesectionheading{\section{Introduction}\label{sec:introduction}}

\IEEEPARstart{T}{his} paper considers the task of salient object detection (SOD), which
aims to detect the most attention-grabbing objects in a scene and then extract pixel-accurate silhouettes for them.
The merit of SOD lies in its many applications, including
foreground map evaluation~\cite{margolin2014evaluate,fan2017structure,Fan2018Enhanced},
visual tracking~\cite{zhang2017online,borji2012adaptive,mahadevan2012connections},
action recognition~\cite{abdulmunem2016saliency}, image retrieval~\cite{he2012mobile,liu2013model},
information discovery~\cite{zhu2015unsupervised,liu2012web}, image contrast enhancement~\cite{gu2015automatic},
person re-identification~\cite{zhao2013unsupervised} image segmentation~\cite{donoser2009saliency,oh2017exploiting},
video segmentation~\cite{fan2019shifting}, photo synthesis~\cite{chen2009sketch2photo},
content-aware image editing~\cite{cheng2010repfinder},
image caption~\cite{ramanishka2017top}, and video compression~\cite{guo2010novel,hadizadeh2014saliency},
style transfer~\cite{liu2019image,TIP20_SP_NPR},
image matching~\cite{toshev2007image},
autonomous underwater robots~\cite{islam2020svam},
camouflaged object detection~\cite{fan2020camouflaged,cheng2022implicit}, aesthetic scoring~\cite{tu2020image}, self-driving vehicles~\cite{kim2019grounding}, plant species identification~\cite{kumar2012leafsnap}, dichotomous image segmentation\footnote{\url{https://xuebinqin.github.io/dis/index.html}}, VR/AR~\cite{qin2021boundary}\footnote{AR CUT \& PASTE: \url{https://www.youtube.com/watch?v=VxJmS8avjbY}.}, Sony's BRAVIA XR TV\footnote{\url{https://www.youtube.com/watch?v=4LnCuTAlVno&feature=youtu.be}.}, \etc.
\begin{figure}[t!]
	\centering
	\begin{overpic}[width=\columnwidth]{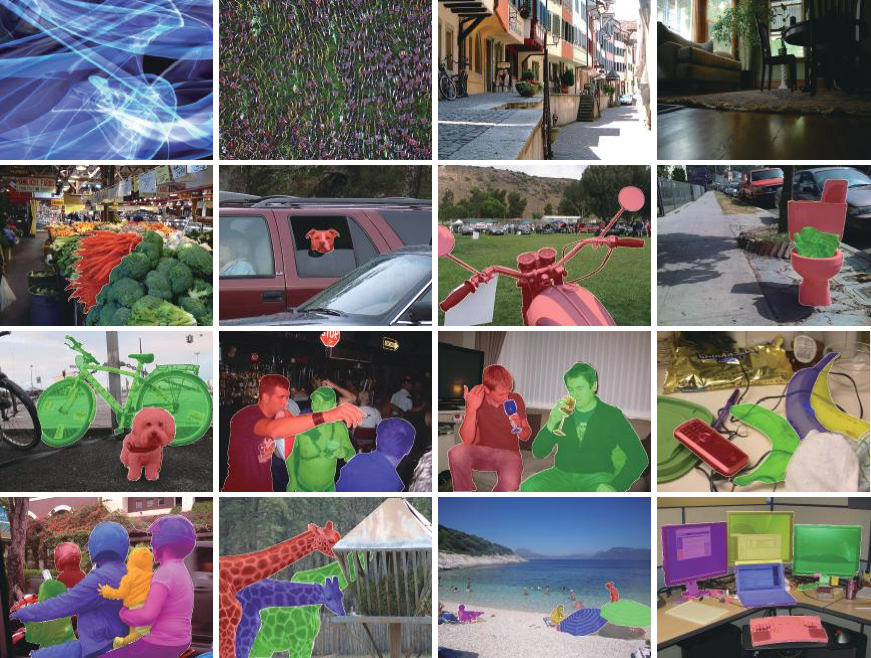}
        \AddText{8}{65}{\scriptsize {non-salient}}
        \AddText{33}{65}{\scriptsize{non-salient}}
        \AddText{58}{65}{\scriptsize {non-salient}}
        \AddText{83}{65}{\scriptsize {non-salient}}

        \AddText{8}{45}{\scriptsize {carrot}}
        \AddAttr{1}{53}{\scriptsize {Attr:HO,OC,SC,SO}}
        \AddText{33}{45}{\scriptsize {dog}}
        \AddAttr{26}{53}{\scriptsize {Attr:HO,OC,SC,SO}}
        \AddText{58}{45}{\scriptsize {motorcycle}}
        \AddAttr{51}{53}{\scriptsize {Attr:HO,SC}}
        \AddText{90}{40}{\scriptsize {toilet}}
        \AddAttr{76}{53}{\scriptsize {Attr:HO,OC,AC}}
        \AddText{88}{48}{\scriptsize{book}}

        \AddText{15}{30}{\scriptsize {bicycle}}
        \AddAttr{1}{34}{\scriptsize {Attr:OC,SC}}
        \AddText{15}{23}{\scriptsize {dog}}
        \AddText{32}{23}{\scriptsize {person}}
        \AddAttr{26}{34}{\scriptsize {Attr:HO,OC,OV,SC}}
        \AddText{60}{26}{\scriptsize {wine glass}}
        \AddText{58}{23}{\scriptsize {person}}
        \AddAttr{51}{34}{\scriptsize {Attr:OV}}
        \AddText{79}{20}{\scriptsize {cell phone}}
        \AddText{88}{26}{\scriptsize {banana}}
        \AddAttr{76}{34}{\scriptsize {Attr:HO,OC,SC}}

        \AddText{5}{8}{\scriptsize {person}}
        \AddAttr{1}{15}{\scriptsize {Attr:BO,OC}}
        \AddText{30}{8}{\scriptsize {giraffe}}
        \AddAttr{26}{15}{\scriptsize {Attr:HO,OC,OV}}
        \AddText{54}{3}{\scriptsize {person}}
        \AddAttr{51}{15}{\scriptsize {Attr:HO,OC,AC}}
        \AddText{62}{1}{\scriptsize {umbrella}}
        \AddText{83}{12}{\scriptsize {computers}}
        \AddAttr{76}{15}{\scriptsize {Attr:HO}}
        \AddText{89}{7}{\scriptsize {laptop}}
        \AddText{80}{3}{\scriptsize {keyboard}}
    \end{overpic}
    \vspace{-15pt}
	\caption{Examples from our new SOC dataset, including
    \emph{non-salient} (first row) and \emph{salient} object images (rows 2 to 4).
    For salient object images, an instance-level ground-truth map (different color),
    object attributes (Attr) and category labels are provided.}
    \label{fig:DatasetExample}
\end{figure}
However, existing SOD datasets
~\cite{LiuSZTS07Learn,alpert2007image,2001iccvSOD,ChengPAMI15,borji2012salient,yan2013hierarchical,YangZLRY13Manifold,li2014secrets,li2015visual,wang2017learning,yang2021biconnet}
are flawed either in their data collection procedure or data quality. Specifically, most datasets assume that an image should contain at least \textit{one salient object, and thus they discard images that do not contain any salient objects.} We call this \emph{\textbf{data selection bias~\cite{torralba2011unbiased}.}}

Moreover, existing datasets typically contain images with a single object or several uncluttered objects.
These datasets do not adequately reflect the complexity of real-world images, where
scenes usually contain multiple objects amidst significant clutter.
As a result, all top-performing models trained on the existing large-scale datasets (\eg, DUTS~\cite{wang2017learning}) have nearly saturated performance (\eg, SCRN~\cite{wu2019stacked} has an $S\textrm{-}measure > 0.9$ on ECC~\cite{yan2013hierarchical}), but still achieve unsatisfactory results on realistic images (\eg, $S\textrm{-}measure < 0.8$ on \ourdataset~\cite{fan2018salient}).
As the current SOD models are biased towards ideal conditions, their effectiveness may be
impaired once they are applied to real-world scenes.
To solve this problem, it is important to introduce a dataset with more realistic conditions.

\begin{figure*}[t!]
  \centering
  \includegraphics[width=\linewidth]{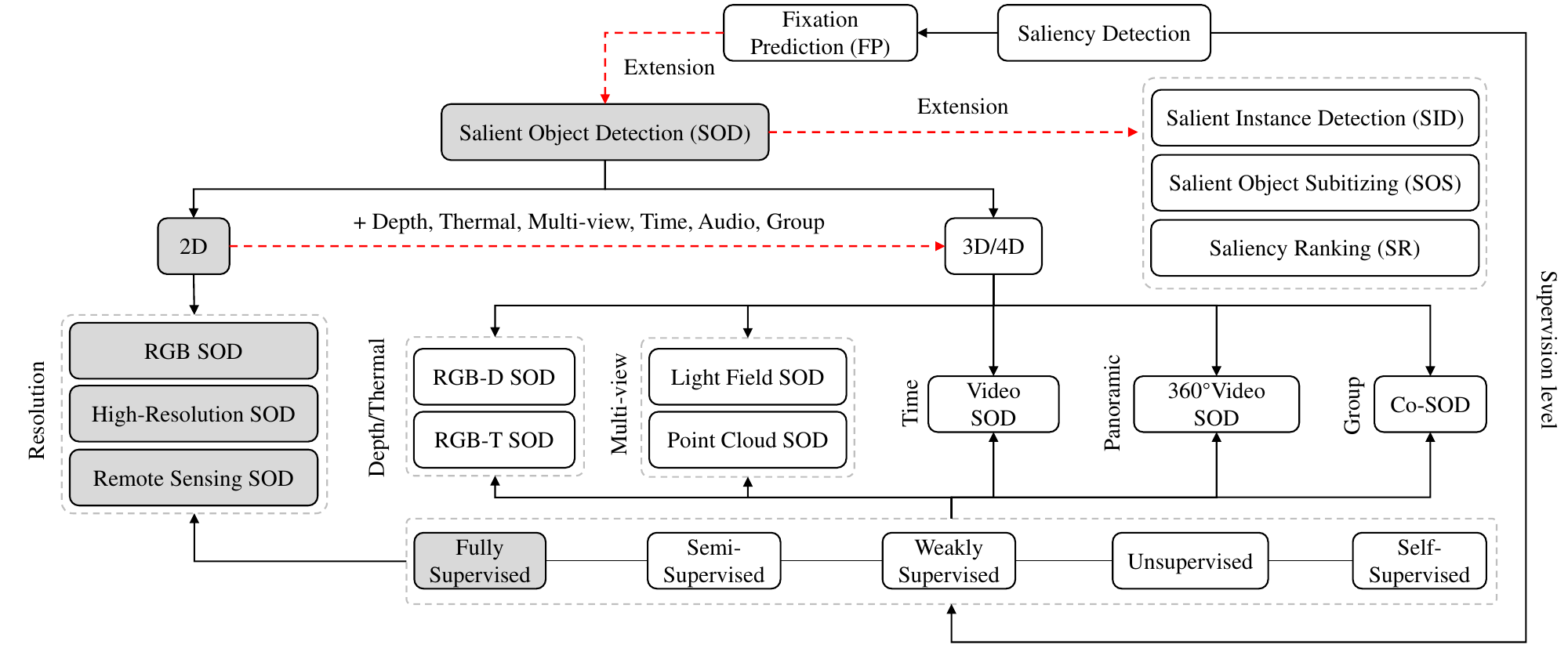}
  \vspace{-10pt}
  \caption{Taxonomy of the saliency detection task. We highlight the scope of this study in gray. See \secref{sec:relatedworks} for details.
  }\label{fig:Scope}
  \vspace{-10pt}
\end{figure*}

Another issue faced by the RGB SOD community is that only the overall performance of the models
can be analyzed using existing datasets. This is because none of the datasets contain attributes that
reflect different challenges.
Having such attributes would help
i) provide deeper insight into the SOD problem,
ii) enable the pros and cons of the SOD models to be investigated, and
iii) allow the model performances to be objectively assessed
from different perspectives. Finally, with a given saliency encoder, \eg, the backbone network, existing saliency models are designed to achieve mapping from the training image set to the training ground-truth set. We thus argue that efforts on improving the dataset, \eg, fixing the data bias issue, can yield higher performance gains than focusing only on the decoder design.
Towards this, we investigate several dataset-enhancement strategies, including
label smoothing to highlight salient boundaries, random image augmentation to adapt saliency models to various scenarios, and self-supervised learning as a form of regularization to learn from small datasets. Extensive experiments validate the effectiveness of these tricks.

Our contributions are summarized as follows:

\begin{itemize}
\item [1)] \textbf{Dataset.}
We collect a new high-quality SOD dataset, named ``Salient Objects in Clutter,'' or \textbf{SOC}.
\ourdataset~is the largest instance-level SOD dataset to date, containing 6,000 images from more than 80 common categories.
It differs from existing datasets in three aspects:
i) Salient objects have category annotations, which can be used for
new research problems, such as weakly supervised SOD.
ii) The inclusion of non-salient images and objects makes this dataset more realistic and challenging than the existing ones.
iii) Salient objects have attributes that reflect various situations encountered in the real world, such as \textit{motion blur}, \textit{occlusion} and \textit{background clutter}.
As a consequence, \ourdataset~narrows the gap between existing datasets and real-world scenes.

\item [2)] \textbf{Review \& Benchmark.}
We present the largest scale RGB SOD study, reviewing \textbf{203} representative models including \textbf{84} algorithms using handcrafted features and \textbf{119} deep learning based models.
Besides, we also maintain an online benchmark (\ie, \url{https://github.com/DengPingFan/SODBenchmark}.) to dynamically trace the development of this field.
In addition, we provide the most comprehensive benchmark of the \textbf{100 representative} SOD models.
To evaluate the models, for the first time, we not only present the overall but also an attribute performance evaluation. This allows a deeper understanding of the models and provides a more complete benchmark.

\item [3)] \textbf{Strategy.} We investigate the biased dataset issue and introduce three dataset-enhancement strategies; namely, label smoothing to make the model aware of the salient boundaries, random image augmentation to adapt the saliency models to various common scenarios, and self-supervised learning as a regularization technique to learn from small datasets.
Despite the apparent simplicity of our strategies, we can achieve an average absolute improvement of 1.14\% $S_\alpha$ over nine existing cutting-edge models.

\item [4)] \textbf{Discussions \& Future Directions.}
Based on our \ourdataset, we present the pros and cons of the current SOD algorithms, discuss
several under-investigated open issues, and provide potential future directions at six levels, \eg, the dataset level, task level, model level, supervision level, evaluation level, and application level.
\end{itemize}

This work extends our previous conference version~\cite{fan2018salient} in the following aspects.
First, we provide more details on our \ourdataset, including sample images without salient objects, images with attributes, and statistics of the attributes. Second, we study three novel training dataset related strategies to fully utilize the non-salient object data and achieve the new state-of-the-art performance. Third, we conduct the largest-scale (46 traditional and 54 deep learning models) benchmarking of SOD models on our \ourdataset.
Finally, based on our benchmarking results, we highlight several fundamental research directions and challenges in the SOD.

\begin{table*}[t!]
  \centering
  \scriptsize
  \renewcommand{\arraystretch}{0.2}
  \renewcommand{\tabcolsep}{2.8mm}
  \caption{Summary of popular SOD datasets. Our SOC is the only one meeting all requirements. According to \cite{wang2021salient}, these datasets can be grouped into three types: early ($\blacktriangle$), popular/modern ($\blacklozenge$), and special ($\lozenge$).
  See \secref{sec:SODDataset} for more details.
  }\label{tab:datasetCompare}
  \vspace{-10pt}
  \begin{tabular}{c|r|c|c|cccccccc}
  \toprule
   \#  & Dataset & Year & Publ. & High-Quality & \textbf{$\geq 5k$} & Non-Salient & Attribute & Category & Bounding Box & Object & Instance \\
  \midrule
  1& MSRA-A, -B~\cite{LiuSZTS07Learn}~$\blacktriangle$    & 2007 & CVPR & \checkmark & \checkmark &          -  &   -   &   -    & \checkmark &    \checkmark      &    -   \\
  2& SED1, SED2~\cite{alpert2007image}~$\blacktriangle$          & 2007 & CVPR & \checkmark &     -       &         -   &  -    &    -   &       -     & \checkmark &   -    \\
  3& ASD~\cite{achanta2009frequency}~$\blacktriangle$      & 2009 & CVPR & \checkmark &     -       &        -    &   -   &   -    &   -         & \checkmark &   -    \\
  4& SOD~\cite{movahedi2010design}~$\blacklozenge$       & 2010 &CVPRW & \checkmark &      -      &         -   &    -  &    -   &      -      & \checkmark &   -    \\
  5& M10K~\cite{ChengPAMI15}~$\blacklozenge$         & 2011 & CVPR & \checkmark & \checkmark &   -   &    -   &      -      &     -   & \checkmark &   -     \\
  6& Judd-A~\cite{borji2012salient}~$\blacktriangle$      & 2012 & ECCV & \checkmark &    -        &         -   &   -   &   -    &       -     & \checkmark &   -    \\
  7& DU-O~\cite{YangZLRY13Manifold}~$\blacklozenge$ & 2013 & CVPR & \checkmark & \checkmark &         -   &    -  &    -   & \checkmark & \checkmark &   -    \\
  8& ECC~\cite{yan2013hierarchical}~$\blacklozenge$    & 2013 & CVPR & \checkmark &    -        &         -   &  -    &    -   &        -    & \checkmark &   -    \\
  9& PASCAL-S~\cite{li2014secrets}~$\blacklozenge$       & 2014 & CVPR & \checkmark &       -     &          -  &    -  &     -  &       -     & \checkmark &  -     \\
  10& HKU~\cite{li2015visual}~$\blacklozenge$          & 2015 & CVPR & \checkmark &     -       &         -   &    -  &   -    &      -      & \checkmark &    -   \\
  11& SOS~\cite{zhang2015salient}~$\lozenge$           & 2015 & CVPR & \checkmark &  \checkmark       &         -   &    -  &   -    &  \checkmark   & - &    -   \\
  12& MSO~\cite{zhang2015salient}~$\lozenge$           & 2015 & CVPR & \checkmark &     -       &         -   &    -  &   -    & \checkmark &  - &    -   \\
  13& XPIE~\cite{xia2017and}~$\lozenge$              & 2017 & CVPR & \checkmark & \checkmark &         -   &   -   &    -   &      -      & \checkmark &    -   \\
  14& ILSO~\cite{li2017instance}~$\lozenge$          & 2017 & CVPR &     -       &    -        &        -    &  -    &    -   &       -     & \checkmark & \checkmark \\
  15& JOT~\cite{jiang2017joint}~$\lozenge$          & 2017 & FCS  & \checkmark & \checkmark & \checkmark &  -    &    -   &      -      & \checkmark &   -    \\
  16& DUTS~\cite{wang2017learning}~$\blacklozenge$       & 2017 & CVPR & \checkmark & \checkmark &          -  &   -   &    -   &        -    & \checkmark &   -    \\
  \hline
  17& \textbf{SOC} (\textbf{OUR})~$\blacklozenge$       & 2022  & -- & \checkmark & \checkmark & \checkmark & \checkmark & \checkmark & \checkmark & \checkmark & \checkmark \\
  \bottomrule
  \end{tabular}
\end{table*}

\section{Related Work}\label{sec:relatedworks}
\subsection{Scope}
%ISOD, HRSOD, RGB-D SOD
Salient object detection originated from the task of fixation prediction (FP)~\cite{itti1998model,borji2021saliency}, switching attention regions for accurate object-level regions. SOD can be traced back to the seminal works~\cite{liu2007learning,liu2010learning}. Current algorithms have been developed for 2D images of limited resolution (width or height $<$ 500 pixels), high-resolution (\ie, 1080p, 4K)~\cite{zhang2021look,zeng2019towards} and even remote sensing data~\cite{zhang2020dense}.
According to the supervision strategy, there are five types of SOD models: fully supervised~\cite{zhuge2021salient}, semi-supervised~\cite{zhou2019semisu}, weakly supervised~\cite{li2018weakly,zhang2019capsal, zhang2020weakly}, unsupervised~\cite{zeng2018unsupervised,zhang2018deep,nguyen2019deepusps}, and self-supervised~\cite{wang2019retrieval,nguyen2019deepusps}.

%RGB-T SOD, Light Field SOD, Co-SOD, VSOD, RSSOD, 360SOD.
Recently, several interesting extensions of SOD have also been introduced, such as salient instance detection (SID)~\cite{li2017instance,fan2019s4net}, salient object subitizing (SOS)~\cite{zhang2015salient,islam2018revisiting,he2017delving}, and saliency ranking~\cite{kalash2019relative,siris2020inferring}. A taxonomy of the saliency detection task is shown in \figref{fig:Scope}. Different from previous SOD reviews~\cite{borji2012state,borji2015salient,nguyen2018attentive,borji2019salient,zhang2018review,cong2018review,han2018advanced,borji2018saliency,wang2021salient}, we mainly focus on 2D salient object detection in a fully supervised manner. We highlight the scope of this study in gray.
For other closely related 3D/4D SOD tasks, we refer readers to
recent survey and benchmarking works such as RGB-D SOD~\cite{fan2020rethinking,zhou2021rgb}, Event-RGB SOD (ERSOD)~\footnote{ERSOD: \url{https://github.com/jxr326/ERSOD-Net}.}, Light Field SOD~\cite{jiang2020light},  Co-SOD~\cite{deng2020re},
360$^\circ$Video SOD~\cite{li2019distortion}, and Video SOD ~\cite{fan2019shifting}.

\subsection{SOD Datasets}\label{sec:SODDataset}
In this section, we briefly discuss existing datasets designed for
SOD tasks, focusing in particular on aspects including annotation type, number of salient objects per image, number of images, and image quality. These datasets are listed in \tabref{tab:datasetCompare}.

\begin{figure}[t!]
	\centering
	\begin{overpic}[width=\columnwidth]{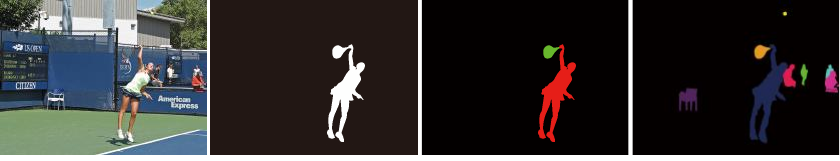}
    \put(5,-4){\footnotesize {(a) Image}}
    \put(29,-4){\footnotesize {(b) Previous}}
    \put(56,-4){\footnotesize {(c) Ours}}
    \put(77,-4){\footnotesize {(d) Segmentation}}
    \end{overpic}
    \vspace{-10pt}
    \caption{Previous SOD datasets only annotate the images
    by drawing pixel-accurate silhouettes around salient objects (b).
    Different from object segmentation datasets~\cite{lin2014microsoft} (d)
    where (objects are not necessarily \textbf{salient}),
    our SOC provides \emph{salient instances} (c).
    We provide a high-quality and large-scale annotated dataset comprised of images
    that better capture the properties of real-world scenes.
    }\label{fig:AnnotationDifference}
\end{figure}
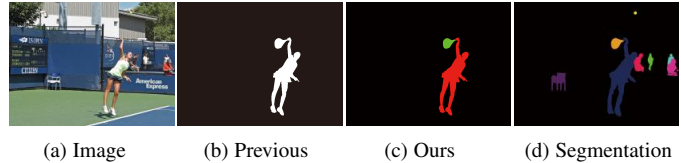

Early datasets are either limited in their numbers of images or
in their coarse annotations of salient objects.
For example, salient objects in the original version of \textbf{MSRA-A}~\cite{LiuSZTS07Learn} and
\textbf{MSRA-B}~\cite{LiuSZTS07Learn} are only roughly annotated in the form of bounding boxes. \textbf{ASD}~\cite{achanta2009frequency}, \textbf{SED1}~\cite{alpert2007image} and \textbf{M10K}~\cite{ChengPAMI15} contain only one salient object in most images, while the \textbf{SED2}~\cite{alpert2007image} dataset provides two objects per image but contains only 100 images.
In order to improve the quality of datasets,
researchers in recent years have started to collect images with multiple objects
in relatively complex and cluttered backgrounds.
The new datasets include \textbf{ECC}~\cite{yan2013hierarchical},
\textbf{DU-O}~\cite{YangZLRY13Manifold},
\textbf{Judd-A}~\cite{borji2012salient},
and \textbf{PASCAL-S}~\cite{li2014secrets}.
These datasets are improved in terms of both annotation quality and number of images, compared to their predecessors.
To resolve the shortcomings still present, some datasets (\eg, \textbf{HKU}~\cite{li2015visual}, \textbf{XPIE}~\cite{xia2017and},
and \textbf{DUTS}~\cite{wang2017learning}) provide
large amounts of pixel-wise labeled images (\figref{fig:AnnotationDifference}.b)
with more than one salient object per image.
However, they ignore non-salient objects (1$^{st}$ row in \figref{fig:DatasetExample}) and
do not offer instance-level annotations (\figref{fig:AnnotationDifference}.c). Jiang~\etal~\cite{jiang2017joint} collected roughly 6K \textit{simple background
images} (most of them are pure texture images) to cover non-salient scenes.
However, their dataset, named \textbf{JOT}, falls short in capturing the complexity of real-world scenes. The dataset of \textbf{ILSO}~\cite{li2017instance} contains instance-level salient object annotations but only roughly labeled boundaries, as shown in \figref{fig:HighQualityAnnotation}. Beyond the ``standard'' SOD datasets, there are also several other special datasets that introduce new tasks, such as salient object subitizing (\ie, \textbf{SOS}~\cite{zhang2015salient} and its subset \textbf{MSO}~\cite{zhang2015salient}).

To sum up, as discussed above, existing datasets mostly focus on images with clear salient objects and simple backgrounds. Considering the aforementioned limitations of existing datasets, a more realistic dataset, containing non-salient objects,
textures ``in the wild'', and salient objects with attributes, is needed for future investigations in this field. Such a dataset could offer deeper insight into the strengths and weaknesses of SOD models, and help overcome performance saturation.
Our \textbf{SOC}~is unique in that it provides various high-quality annotations, as shown in \tabref{tab:datasetCompare}.

\subsection{SOD Models}
We have noticed that, from 1998 to the end of Feb. 2021, more than 10,000 papers on saliency detection or related field have been published. In this section, we try our best to summarize
those published in top conferences (\eg, NeurIPS, CVPR) and journals (\eg, TPAMI, TIP), as well as some high-quality open-access (\ie, arXiv) works. Instead of briefly describing the pipeline of each model, we summarize key components to provide a global view.

As shown in \tabref{tab:HandCraftModelSummary}, a number of different approaches have been designed to tackle SOD using super-pixel, proposal, or  edge/boundary annotations under different levels of supervision, such as unsupervised, semi-supervised, and fully supervised. Using common aggregation strategies (\eg, linear, non-linear), these methods mainly focus on pixels, regions, and patches to design more powerful models. Besides, we note that certain priors (\eg, the center-surround prior, local/global contrast prior, fore/background prior, and boundary prior) are frequently used in these methods.
Some models also utilize different post-processing steps (\eg, conditional random field, morphology, watershed, and max-flow strategies) to further improve the performance.

\begin{table*}[t!]
  \centering
  \scriptsize
  \renewcommand{\arraystretch}{0.2}
  \renewcommand{\tabcolsep}{1.0mm}
  \caption{
  Summary of popular SOD models using handcrafted features.
  \textbf{Agg.:} Aggregation strategy, \eg, LN = linear, NL = non-linear, HI = hierarchical, BA = Bayesian, AD = adaptive, LS = least-square solver, EM = energy minimization, and GMRF = Gaussian MRF.
  \textbf{SL.:} Supervision level, \eg, unsupervised ($\bigstar$), semi-supervised ($\bullet$), weakly supervised ($\leftmoon$), fully supervised ($\circ$), active learning ($A$).
  \textbf{Sp.:} Whether or not superpixel over-segmentation is used.
  \textbf{Pr.:} Whether or not proposal methods are used.
  \textbf{Ed.:} Whether or not edge cues are used.
  \textbf{Post-Pros.:} Whether post-processing methods (\eg, CRF~\cite{Krhenbhl2011EfficientII}, graph-cut~\cite{Boykov2001fast}, GrabCut~\cite{Rother2004grab}, Ncut~\cite{Jianbo2000Nor}), morphology, max-flow (MF)~\cite{boykov2004experimental} or only thresholding are used.
  }\label{tab:HandCraftModelSummary}
  \vspace{-10pt}
\begin{tabular}{c|c|r|c|r|c|c|c|c|c|c|c|c|c}
  \toprule
   & \#  & Model                         & Publ.   & Scholar & Prior.  & Uniqueness    & Component                      & Agg. & SL.        & Sp. & Pr. & Ed. & Post-Pros \\
  \midrule
  \multirow{16}{*}{\vspace{-.4in}\begin{sideways}2010 - 1998\end{sideways}}
  &1&  Itti~\cite{itti1998model}         & TPAMI   & \href{https://scholar.google.com/citations?user=xhUvqK8AAAAJ&hl=zh-CN}{link} & center-surround           & pixel         & Color, Intensity, Orientation     & LN & $\bigstar$ & -  & -  & -  &-\\
  &2&  GBVS~\cite{harel2007graph}        & NeurIPS & \href{https://scholar.google.ae/citations?user=g1L7TtsAAAAJ&hl=zh-CN&oi=sra}{link} & -                         & pixel         & Markovian                         & -  & $\bigstar$ & -  &  - & -  &-\\
  &3&  FT~\cite{achanta2009frequency}    & CVPR    & \href{https://scholar.google.ae/citations?user=lc2HaZwAAAAJ&hl=zh-CN&oi=sra}{link}  & frequency domain          & pixel         & Color, Luminance                  & -  & $\bigstar$ & -  &  - & -  &-\\
  &4&  SR~\cite{hou2007saliency}         & CVPR    & \href{https://scholar.google.com/citations?user=1smFmxAAAAAJ&hl=zh-CN}{link}  & spectral residual         & pixel         & Log Spectrum                      & -  & $\bigstar$ & -  &  - & -  &-\\
  &5&  AIM~\cite{bruce2006saliency}      & NeurIPS & \href{https://scholar.google.ae/citations?user=Gnezf-4AAAAJ&hl=zh-CN&oi=sra}{link}  & maximizing information    & patch         & Shannon's Self-information        & -  & $\bigstar$ & -  &  - & -  &-\\
  &6&  SUN~\cite{zhang2008sun}           & JOV     & \href{https://scholar.google.com/citations?user=wuBvtagAAAAJ&hl=zh-CN&oi=sra}{link}  & self-information          & pixel         & DoG, ICA-derived features         & -  & $\bigstar$ & -  &  - & -  &-\\
\bottomrule
  \end{tabular}
  \vspace{-5pt}
\end{table*}

\begin{table*}[t!]
  \centering
  \scriptsize
  \renewcommand{\arraystretch}{0.2}
  \renewcommand{\tabcolsep}{1.5mm}
  \caption{
  Summary of popular deep learning based SOD models.
  See \tabref{tab:HandCraftModelSummary} for more detailed descriptions.
  MB = MSRA-B dataset~\cite{LiuSZTS07Learn}.
  M10K = MSRA-10K~\cite{ChengPAMI15} dataset.
  P-VOC2010 = PASCAL VOC 2010 semantic segmentation dataset~\cite{pascalvoc2010}.
  CRF = Conditional random fields.
  \textbf{Clicking the scholar will link to the specific author's google scholar.}
  }\label{tab:CNNModelSummary}
  \vspace{-10pt}
\begin{tabular}{c|c|r|c|r|c|c|c|c|c|c|c|c}
  \toprule
   & \#  & Model & Publ. & Scholar & \#Training & Training Dataset & Backbone & SL. & Sp. & Pr. & Ed.& CRF \\
  \midrule
  \multirow{4}{*}{\begin{sideways}2015\end{sideways}}
  &1&  SupCNN~\cite{he2015supercnn}  & IJCV  & \href{https://scholar.google.ae/citations?user=rBWnK8wAAAAJ&hl=zh-CN&oi=sra}{link} & 800                & ECC~\cite{yan2013hierarchical} & - & $\circ$   & \checkmark & - &  -  & - \\
  &2&  LEGS~\cite{wang2015deep}      & CVPR  & \href{https://scholar.google.ae/citations?user=EfTwkXMolscC&hl=zh-CN&oi=sra}{link} & 340+3,000          & PASCAL-S~\cite{li2014secrets}+MB\cite{LiuSZTS07Learn} & - & $\circ$ & - &\checkmark & -  & - \\
  &3&  MDF~\cite{li2015visual}       & CVPR  & \href{https://scholar.google.ae/citations?user=2A2Bx2UAAAAJ&hl=zh-CN&oi=sra}{link} & 2,500              & MB\cite{LiuSZTS07Learn}      &   -      & $\circ$ &\checkmark &  - &\checkmark & -\\
  &4&  MC~\cite{zhao2015saliency}    & CVPR  & \href{https://scholar.google.ae/citations?user=1c9oQNMAAAAJ&hl=zh-CN&oi=sra}{link} & 8,000              & M10K~\cite{ChengPAMI15}      & GoogLeNet~\cite{szegedy2015going} & $\circ$ &\checkmark & -  & - & - \\
  \midrule
  \multirow{12}{*}{\begin{sideways}2016\end{sideways}}
  &5& DSL~\cite{yuan2016dense}      & TCSVT & \href{https://scholar.google.ae/citations?user=QJAr1KoAAAAJ&hl=zh-CN&oi=sra}{link}  & (5,168+10,000)*80\% &    DU-O~\cite{YangZLRY13Manifold}+M10K~\cite{ChengPAMI15} & LeNet~\cite{Lecun1998Grad}/VGGNet16 & $\circ$ & \checkmark & - & - & - \\
  &6&  DISC~\cite{chen2016disc}      & TNNLS & \href{https://scholar.google.ae/citations?user=aNvkklMAAAAJ&hl=zh-CN&oi=sra}{link} & 9,000               & M10K~\cite{ChengPAMI15}      &   -    & $\circ$ &\checkmark& - & - & - \\
\bottomrule
  \end{tabular}
\end{table*}

\begin{table*}[t!]
  \centering
  \scriptsize
  \renewcommand{\arraystretch}{0.1}
  \renewcommand{\tabcolsep}{2.3mm}
  \caption{
  Summary of popular deep learning based SOD models.
  See Tables \ref{tab:HandCraftModelSummary} \& \ref{tab:CNNModelSummary} for more detailed descriptions.
  }\label{tab:CNNModelSummary2}
  \vspace{-10pt}
  \begin{tabular}{c|c|r|c|r|c|c|c|c|c|c|c|c}
  \toprule
   & \#  & Model & Publ. & Scholar & \#Training & Training Dataset & Backbone & SL. & Sp. & Pr. & Ed.& CRF \\
  \midrule
  \multirow{16}{*}{\vspace{-1in}\begin{sideways} 2020 \end{sideways}}
  &93& SUCA~\cite{li2020stacked} & TMM & \href{https://scholar.google.com/citations?user=2Pyf20IAAAAJ&hl=zh-CN&oi=sra}{link} & 10,553 & DUTS~\cite{wang2017learning} & ResNet50      & $\circ$ & - & - & - & - \\
  &94& MIJR~\cite{ma2020salient} & TMM & \href{https://scholar.google.com/citations?hl=zh-CN&user=SGjgjBUAAAAJ}{link}  & 2,500+5,000              & MB~\cite{LiuSZTS07Learn}+DUTS~\cite{wang2017learning} & VGGNet16      & $\circ$ & \checkmark & \checkmark & - & \checkmark\\
  &95& CAGVgg~\cite{mohammadi2020cagnet} & PR & \href{https://scholar.google.com/citations?user=SQ_rJ98AAAAJ&hl=zh-CN&oi=sra}{link} & 10,553 & DUTS~\cite{wang2017learning} & VGGNet/ResNet/NASNet~\cite{zoph2018learning}      & $\circ$ & - & - & - & - \\
  &96& U2Net~\cite{qin2020u2} & PR & \href{https://scholar.google.ae/citations?user=MMLkNtYAAAAJ&hl=zh-CN&oi=sra}{link} & 10,553  & DUTS~\cite{wang2017learning} &  UNet      & $\circ$ & - & - & - & - \\
  &97& SalGAN~\cite{wang2020saliencygan} & TII & \href{https://scholar.google.ae/citations?user=lwV4KfwAAAAJ&hl=zh-CN&oi=sra}{link} & 10,000 & M10K~\cite{ChengPAMI15} & VGGNet16      & $\circ$ & - & - & - & - \\
  &98& ADA~\cite{song2020multi} & AAAI & \href{https://scholar.google.ae/citations?user=JnQts0kAAAAJ&hl=zh-CN&oi=sra}{link}& 2,500+780  & MB~\cite{LiuSZTS07Learn}+NIR~\cite{song2020multi} & VGGNet16      & $\circ$ & - & - & - & - \\
  &99& PFPNet~\cite{wang2020progressive} & AAAI & \href{https://scholar.google.com/citations?hl=zh-CN&user=yryOvLwAAAAJ}{link} & 10,553 & DUTS~\cite{wang2017learning} & ResNet101      & $\circ$ & - & - & - & - \\
\bottomrule
  \end{tabular}
\end{table*}

More recently, many deep learning SOD models based on different network architectures, such as multi-layer perceptrons, fully convolutional networks (FCNs), hybrid networks and capsules, have been proposed and achieve higher performance than traditional methods. According to the learning paradigm, most deep SOD models can be roughly split into two types: single-task learning and multi-task learning methods. We summarize the training data, backbones, and other components in Tables \ref{tab:CNNModelSummary} and \ref{tab:CNNModelSummary2}.

We mainly focus on macro-level statistics rather than micro-level descriptions. We kindly refer readers to the recent architecture review~\cite{wang2021salient}. We hope this comprehensive review can serve as guidance\footnote{Research group: \url{https://github.com/DengPingFan/Saliency-Authors}.} for future researchers in this fast-growing field.

\subsection{Dataset-Enhancement Strategies for Deep Models}
Existing deep SOD models focus on designing effective decoders
\cite{wu2019stacked,wei2020f3net,wang2020deep,li2020complementarity,li2020depthwise,zhang2020dense,liu2021dna} to aggregate features from different levels
of the backbone network \cite{he2016deep,Xie_2017_CVPR,pami21Res2net}.
We argue that, as they employ a mapping function from the input training
image set to the output training ground-truth set,
deep models should also focus on dataset-enhancement strategies
to improve model generalization ability.
Three different strategies have been widely studied,
including label smoothing \cite{rethink_inception},
image augmentation \cite{hide_and_seek,mixup},
and self-supervised learning~\cite{feng2019self}.

Instead of training directly with one-hot supervision, \enquote{label smoothing} techniques learn from smoothed supervision, and can thus relax the supervision signals using the generated smoothing labels \cite{rethink_inception} or disturbed labels \cite{disturblabel}.
Miyato \etal~\cite{distribution_smooth} applied local perturbations to data points to increase the smoothness of the model distribution.
Thulasidasan \etal~\cite{On_Mixup_Training_Nips19} discovered that mix-up training \cite{mixup} with label smoothing can significantly improve model calibration.
To obtain a more robust and generative model, Xie \etal~\cite{disturblabel} randomly replaced a portion of labels with incorrect values in each iteration.
In addition, Wager~\etal~\cite{Wager:2014:ATS:2968826.2968838} demonstrated that corrupting training examples with noise from known distributions within the exponential family can inject appropriate generative assumptions into discriminative models, thus reducing generalization errors.  Peterso \etal~presented a soft-label dataset (CIFAR10H~\cite{human_uncertainty_robust_iccv}) aiming at reflecting human perceptual uncertainty by providing label distributions across categories instead of hard one-hot labels.

Image augmentation \cite{hide_and_seek} is an effective technique for extending the diversity of a training dataset, thus improving model generalization ability. Existing data augmentation techniques can be roughly divided into two categories: 1) human-designed policies, \eg, rotation or scale transformation, and 2) learned policies~\cite{li2020dada,AutoAugment}. For the former, a predefined data augmentation policy is applied to the dataset. Beside the widely used rotation and scale transformations, other extensively studied methods in this category are erasing techniques~\cite{wei2017object,zhong2020random}, which achieve data augmentation by randomly erasing part of the image patch.
Further, mix-up methods \cite{chang2020mixup, guo2019mixup} utilize the mix-up data augmentation strategy to generate new samples from an existing training dataset to mitigate the uncertainty in prediction.
For the latter \cite{li2020dada}, the network learns an image-conditioned data augmentation policy, which is usually parameterized by a deep neural network. In this way, the input image is fed to the data augmentation network to generate augmented samples with hyperparameters that control the degree of data augmentation.

Self-supervised learning \cite{unsupervised_self_supervised_rotation, feng2019self}, also termed as consistency learning, defines an annotation-free pretext task to provide a surrogate supervision signal for feature learning. Conventionally, self-supervised learning is used for unsupervised representation learning to learn the feature embedding of the image or video. Recently, works have defined self-supervised learning as an  auxiliary task, and used it within a weakly supervised \cite{yu2021structure} or semi-supervised learning framework \cite{zhai2019s4l}. Several recent and  representative arts can be found in \cite{chen2021empirical,wang2017transitive,he2020momentum}.

As far as we know, no existing salient object detection works have focused on exploring the dataset bias issue with dataset-enhancement strategies. In this paper, we claim that efforts on developing dataset-improvement strategies can also yield significant performance gains. Further, these solutions are general and can be easily applied to existing saliency detection networks.

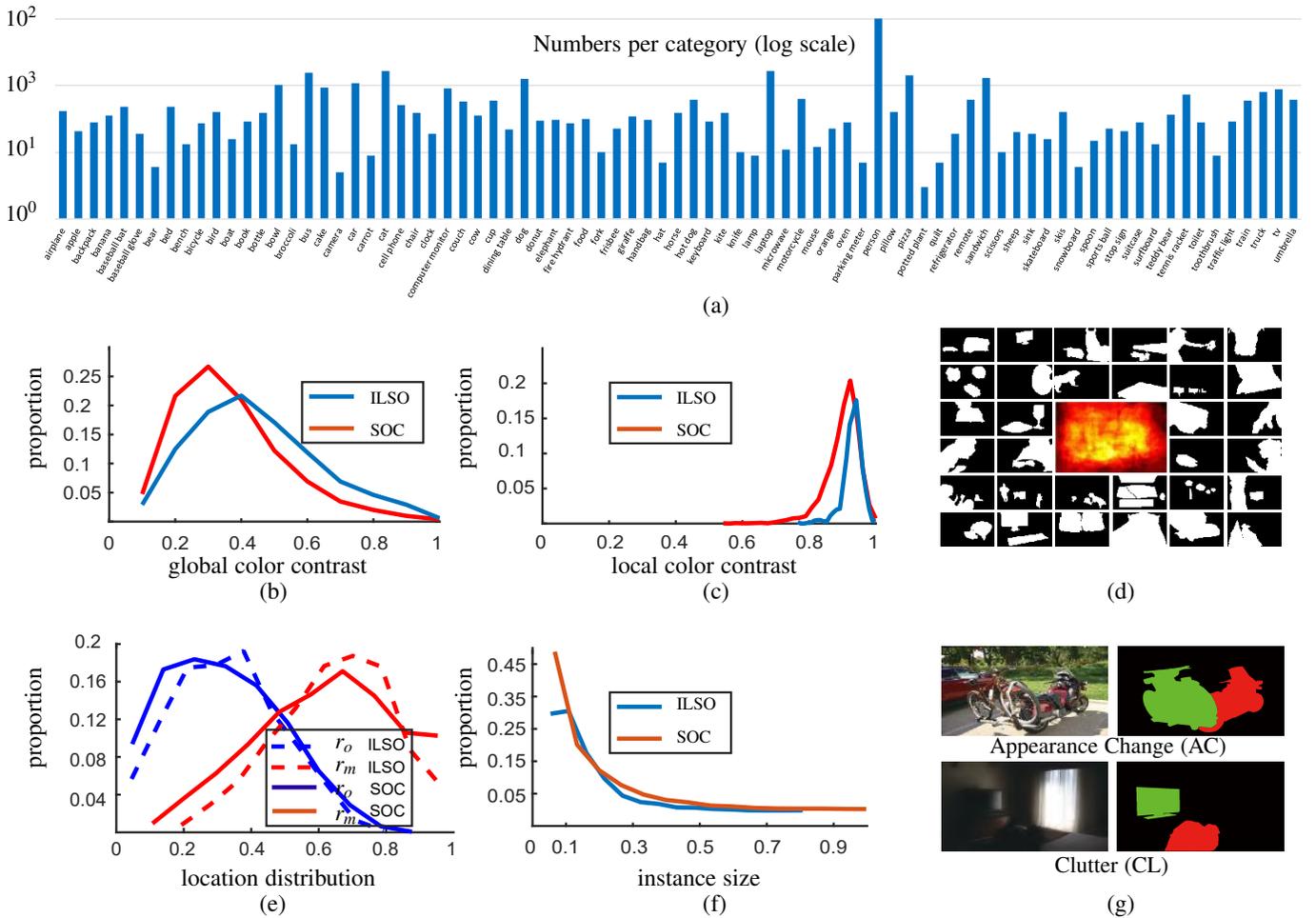
\begin{figure*}[t!]
	\centering
	\begin{overpic}[width=\textwidth]{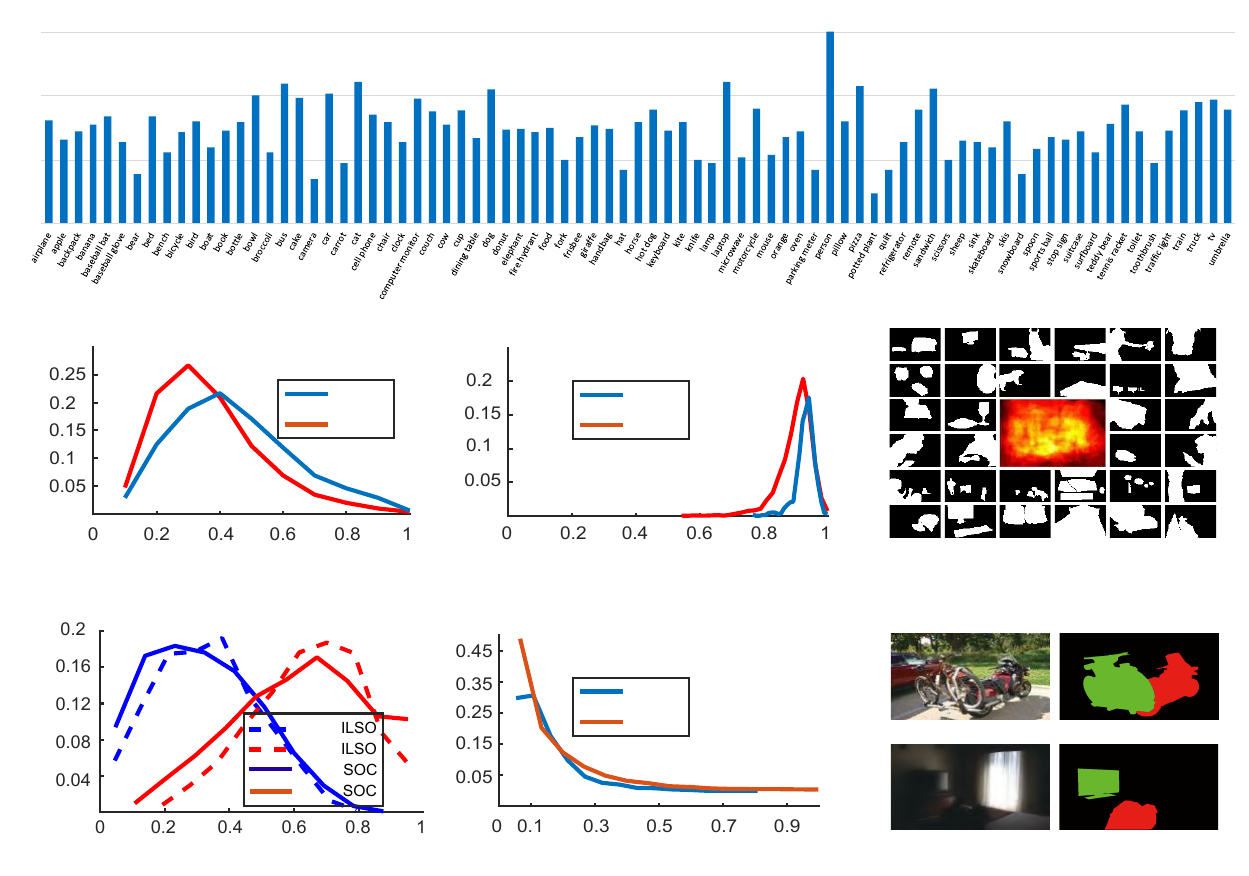}
    \put (40,65){Numbers per category (log scale)}
    \put (-0.5,67){\small {$10^2$} }
    \put (-0.5,62){\small {$10^3$} }
    \put (-0.5,57){\small {$10^1$} }
    \put (-0.5,52){\small {$10^0$} }
    \put(53,45){(a)}

    \put(19,23){(b)}
    \put(27.5,38){\scriptsize {ILSO}}
    \put(27.5,35.5){\scriptsize  {SOC}}
    \put(1,33){\rotatebox{90}{proportion}}
    \put(12,25){global color contrast}

    \put(53,23){(c)}
    \put(51,38){\scriptsize {ILSO}}
    \put(51,35.5){\scriptsize  {SOC}}
    \put(34,33){\rotatebox{90}{proportion}}
    \put(46,25){local color contrast}

    \put(84,23){(d)}

    \put(19,-1){(e)}
    \put(13,1){location distribution}
    \put(1,10){\rotatebox{90}{proportion}}
    \put(24.8,11.5){\small {$r_{o}$}}
    \put(24.8,9.8){\small {$r_{m}$}}
    \put(24.8,8){\small {$r_{o}$}}
    \put(24.8,6){\small {$r_{m}$}}

    \put(84,-1){(g)}
    \put(75,11.2){\small Appearance Change (AC)}
    \put(80,2){\small Clutter (CL)}

    \put(53,-1){(f)}
    \put(51,14.5){\scriptsize {ILSO}}
    \put(51,12){\scriptsize  {SOC}}
    \put(48,1){instance size}
    \put(34,10){\rotatebox{90}{proportion}}
    \end{overpic}
    \vspace{-10pt}
    \caption{(a) Number of annotated instances per category in our SOC dataset.
    (b, c) Global and local color contrast statistics, respectively.
    (d) A set of saliency maps from our dataset and their overlay map. (e) Location distribution of the
    salient objects in SOC. (f) Distribution of instance sizes in the SOC and ILSO~\cite{li2017instance} datasets. (g) Visual examples of attributes.
    Best view on screen and zoomed-in for details.
    }\label{fig:NewStatistical}
\end{figure*}

\section{SOC Dataset}\label{sec:SOCDataset}
\begin{figure}[t!]
\begin{center}
    \includegraphics[width=.98\columnwidth]{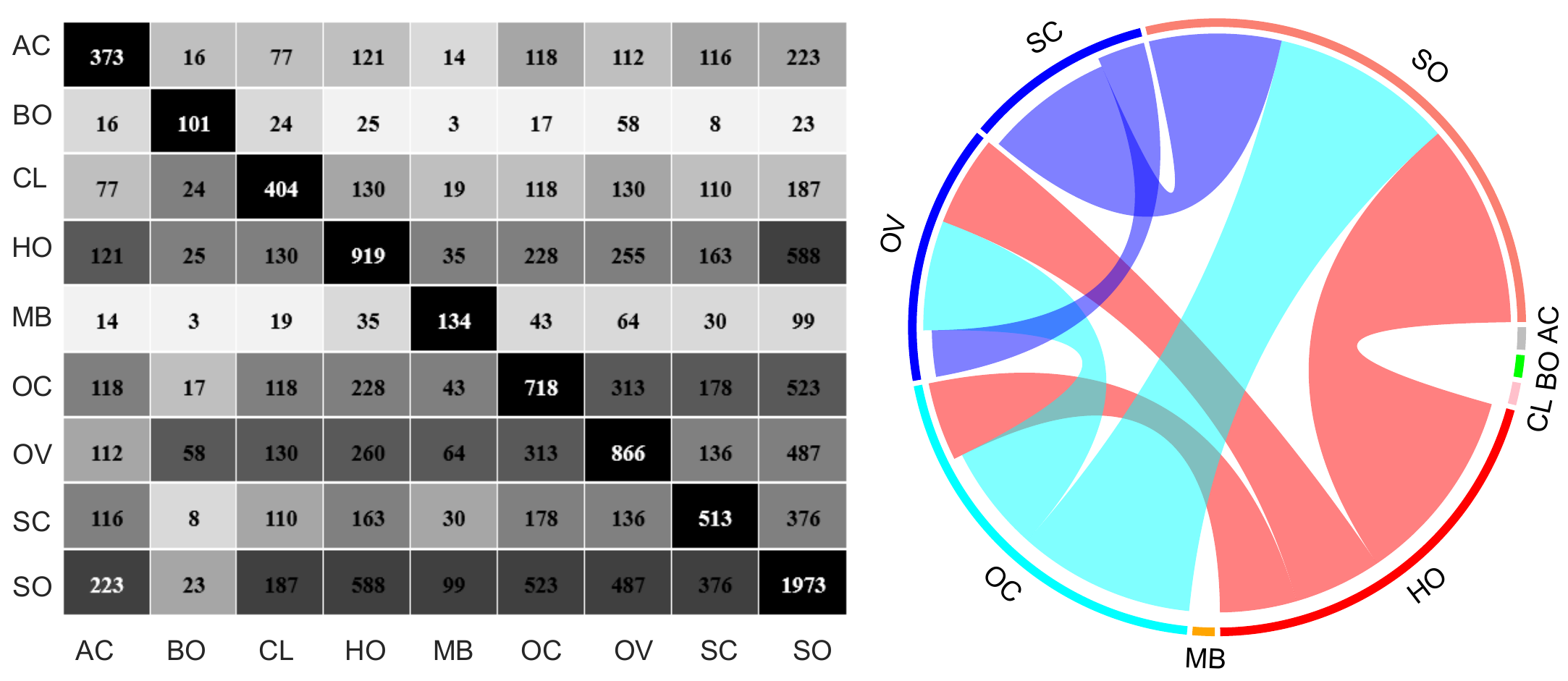}
     \vspace{-10pt}
     \caption{Left: Attribute distribution over salient object images in our SOC dataset.
     Each number in the grid indicates the number of occurrences.
     Right: The dominant dependencies among attributes based on the frequency of occurrences.
     A larger link width indicates a higher probability of an attribute occurring with other ones.}\label{fig:Attributes_distribution}
\end{center}
\end{figure}

In this section, we present details of our new challenging \ourdataset~dataset.
Sample images from \ourdataset~are shown in \figref{fig:DatasetExample}, while statistics regarding the categories and attributes are shown in \figref{fig:NewStatistical} (a) and \figref{fig:Attributes_distribution}, respectively.
Based on the strengths and weaknesses of existing datasets, we identify seven crucial requirements that a comprehensive and balanced dataset should fulfill.

\begin{figure}[b!]
  \centering
  \begin{overpic}[width=\columnwidth]{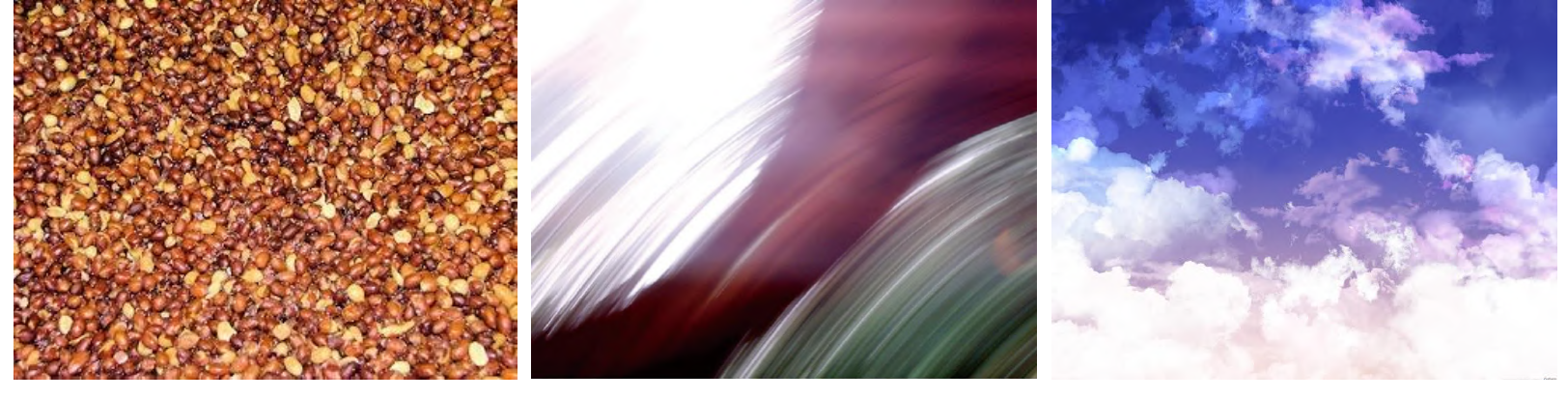}
  \put(14,-2){\small {(a)}}
  \put(46,-2){\footnotesize {(b)}}
  \put(83,-2){\footnotesize {(c)}}
  \end{overpic}
  \caption{Examples of non-salient objects in our dataset. a) Crowded scene, b) motion blur, and c) background with non-interesting regions. }\label{fig:nonsal}
\end{figure}

\textbf{1) Presence of Non-Salient Objects.}
Most existing SOD datasets assume that an image should contain at least one salient object and thus discard images without salient objects~\cite{jiang2017joint}.
However, this assumption is only true under ideal settings, which leads to \emph{data selection bias}.
In realistic settings, images do not always contain salient objects.
For example, some images of amorphous backgrounds, such as sky, grass or textures, contain no salient objects at all~\cite{caesar2016coco}.
The non-salient objects or background ``stuff'' may occupy the entire scene,
and hence heavily constrain the possible locations of a salient object.
Xia \etal~\cite{xia2017and} proposed a \sArt~SOD model
the determines what is or is not a salient object,
indicating that non-salient objects are crucial
for reasoning salient objects.
This suggests that non-salient objects deserve
equal attention in SOD.
Incorporating images containing non-salient objects makes
a dataset more realistic, and hence more challenging.
We define ``\emph{non-salient objects}''
as images without salient objects or images with ``stuff'' categories.
As suggested in~\cite{xia2017and,caesar2016coco},
the ``stuff'' categories include (a) densely distributed similar objects,
(b) fuzzy shapes, and (c) regions without semantics,
as illustrated in \figref{fig:nonsal} (a)-(c), respectively.

To avoid data selection bias, we selected images randomly and automatically, as suggested by Torralba and Efros~\cite{torralba2011unbiased}. Based on the characteristics of non-salient objects, we randomly collected
783 texture images from the DTD~\cite{lazebnik2005sparse} dataset.
To enrich the diversity, 2,217 images including aurora, sky,
crowds, store and many other kinds of realistic scenes were gathered from the
Internet and other datasets~\cite{lin2014microsoft,li2014secrets,2001iccvSOD,jiang2017joint}.

\textbf{2) Number and Categories of Images.}
Providing a large number of images is essential for capturing the diversity
and abundance of real-world scenes.
Moreover, with large amounts of data, SOD models can avoid overfitting and enhance generalization.
To this end, we first randomly gathered 3,000 images from the MS-COCO dataset~\cite{lin2014microsoft},
which contains `everyday scenes of common objects in their \textbf{natural} context.'
Then, more than 80 object categories (see \supp{supplementary materials}) were annotated. Note that we separated the process
of data selection and labeling to avoid data selection bias, as discussed in~\cite{torralba2011unbiased}.
Please refer to the subsection ``\textit{7) High-Quality Salient Object Labeling}'' for details on this. \figref{fig:NewStatistical} (a) shows the number of salient objects in each category. As can be seen, the ``person'' category accounts for a large proportion of the data, which is reasonable as people usually appear in daily scenes along with other objects.
We divided our dataset (3k non-salient images and 3k salient images) into training,
validation and test sets in the ratio of 6:2:2.

\textbf{3) Global vs. Local Color Contrast of Salient Objects.}
As described in~\cite{li2014secrets},
the term ``salient'' is related to the global/local contrast of the foreground
and background. It is essential to confirm whether the salient objects
are easy to detect. For each object, we compute separate RGB color histograms for
the foreground and background. Then, $\chi^2$ distance is utilized to
measure the distance between the two histograms.
The global and local color contrast distributions
are shown in \figref{fig:NewStatistical} (b) and (c), respectively.
Compared to ILSO, the SOC dataset has a higher proportion of objects with low global
and local color contrast.

\begin{table*}[t!]
\scriptsize
\caption{List of salient object image attributes and their corresponding descriptions.
These attributes are derived by studying the characteristics of existing datasets.
Some visual examples can be found in \figref{fig:DatasetExample} and \figref{fig:NewStatistical} (g).
For more examples, please refer to the supplementary materials.
}\label{tab:Attr}
\vspace{-10pt}
\begin{tabular*}{\linewidth}{ll}
  \toprule
  Attribute & Description \\
  \midrule
  AC \emph{\textbf{(Appearance Change)}} &  Obvious illumination change in the object region.\\
  BO  \emph{\textbf{(Big Object)}} &  The ratio between the object area and the image area is larger than 0.5.\\
  CL \emph{\textbf{(Clutter)}}  &  Foreground and background regions around the object have similar colors. We labeled images with a global \\ & color contrast
       value larger than 0.2 and local color contrast value smaller than 0.9 as cluttered images (\secref{sec:SOCDataset}).\\
  HO \emph{\textbf{(Heterogeneous Objects) \ \ }}  &  Objects composed of visually distinctive/dissimilar parts.\\
  MB  \emph{\textbf{(Motion Blur)}} &  Objects have fuzzy boundaries due to camera shaking or motion.\\
  OC  \emph{\textbf{(Occlusion)}} &  Objects are partially or fully occluded.\\
  OV \emph{\textbf{(Out-of-View)}}  &  Part of the object is clipped by the image boundaries.\\
  SC  \emph{\textbf{(Shape Complexity)}} &  Objects have complex boundaries, such as thin parts (\eg, the foot of animal) and holes.\\
  SO \emph{\textbf{(Small Object)}}  &  The ratio between the object area and the image area is smaller than 0.1.\\
  \bottomrule
  \end{tabular*}
\end{table*}

\textbf{4) Locations.}
\emph{Center bias} has been identified as one of the most significant and challenging
biases pertaining to saliency detection datasets
\cite{borji2015salient,li2014secrets,Judd_2012}.
\figref{fig:NewStatistical} (d) illustrates a set of images and their overlay
map (\ie, average mask map).
As can be seen, although salient objects are located at different positions,
the overlay map shows that somehow these images are still center biased.
Unfortunately, previous benchmarks have often adopted this incorrect approach to analyze the positional distribution of salient objects.
To avoid this misleading phenomenon, in \figref{fig:NewStatistical} (e), we plot the statistics of two quantities, $r_o$ and $r_m$, which denote how far an object center
and its farthest (margin) point are from the image center, respectively.
Both $r_o$ and $r_m$ are divided by half the diagonal length of the image
for normalization, such that $r_o, r_m \in [0, 1]$.
From these statistics, we observe that the salient objects in our dataset do not suffer from center bias.

\textbf{5) Size of Salient Objects.}
The size of an instance-level salient object
is defined as the proportion of its pixels to those in the overall image~\cite{li2014secrets}.
As shown in \figref{fig:NewStatistical} (f), the sizes of salient objects
in our SOC vary greatly compared with
the only other existing instance-level dataset, ILSO~\cite{li2017instance}.
Further, there is a higher proportion of medium-sized objects in SOC.

\textbf{6) Salient Objects with Attributes.}
Having attribute information for the images in a dataset helps
objectively assess the performance of models
over different types of parameters and variations.
It also allows for the inspection of model failures.
To this end, we define a
set of attributes to represent specific situations
encountered in common scenes,
such as \emph{motion blur},
\emph{occlusion} and \emph{cluttered background}
(summarized in \tabref{tab:Attr}).
Note that an image can be annotated with multiple attributes as these attributes are not exclusive.

Inspired by~\cite{perazzi2016benchmark}, we present
the distribution of attributes over the dataset in the left of  \figref{fig:Attributes_distribution}.
The \emph{SO} attribute makes up the largest proportion due to our accurate instance-level annotations
(\eg, the tennis racket in \figref{fig:AnnotationDifference}).
The \emph{HO} attribute also accounts for a large proportion, because the real-world scenes are composed of different constituent materials.
\emph{Motion blur (MB)} is more common in video frames, but also sometimes occurs in still images. Thus, \emph{MB} images make up a relatively small proportion of our dataset.
Since a realistic image usually contains multiple attributes, we show
the dominant dependencies among attributes based on the frequency of occurrence on the right of \figref{fig:Attributes_distribution}.
For example, a scene containing several heterogeneous objects is likely to
have a large number of objects occluding each other and forming complex spatial structures.
Thus, the \emph{HO} attribute has a strong dependency with \emph{OC}, \emph{OV}, and \emph{SO}.

\begin{figure}[t!]
  \centering
  \begin{overpic}[width=.98\columnwidth]{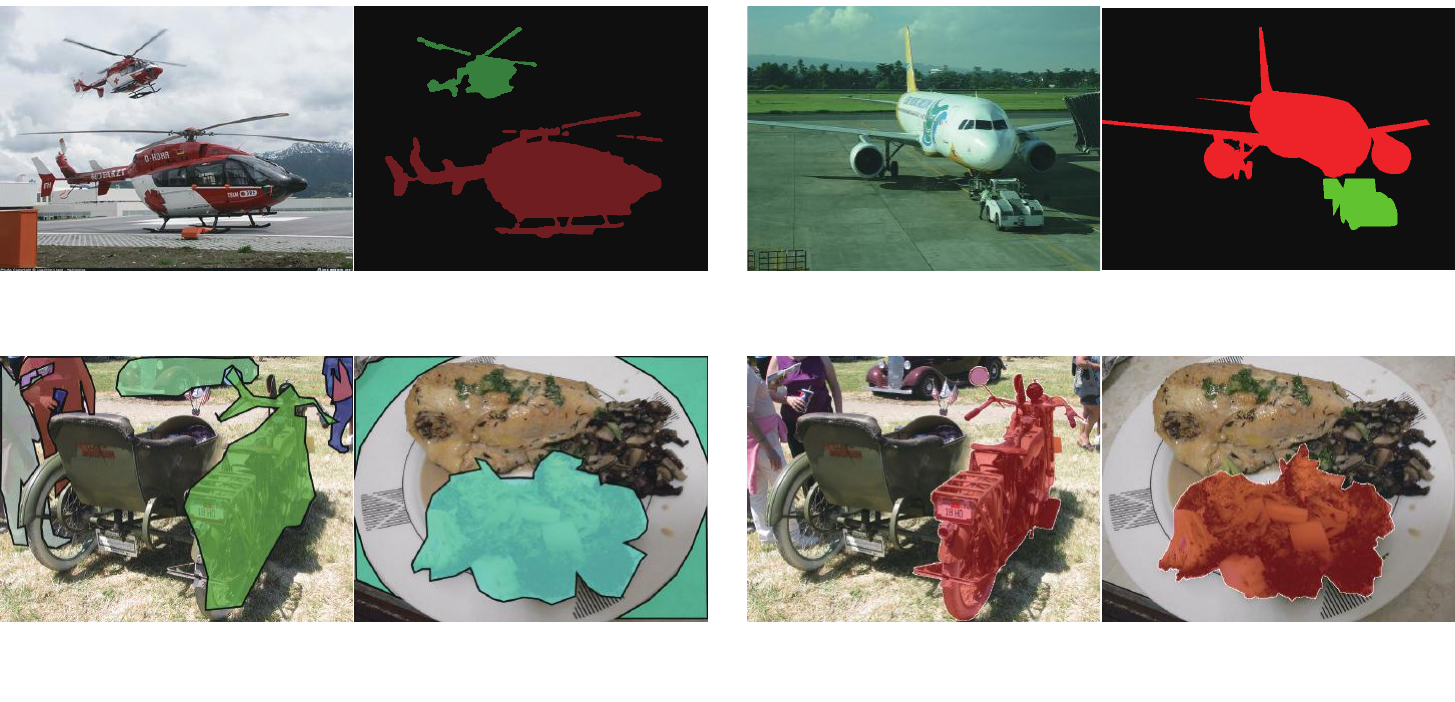}
  \put(16,26){\footnotesize {(a) ILSO} }
  \put(68,26){\footnotesize {(b) SOC} }
  \put(13,2){\footnotesize {(c) MS-COCO} }
  \put(68,2){\footnotesize {(d) SOC} }
  \end{overpic}
  \vspace{-15pt}
  \caption{Compared with the recent instance-level ILSO dataset~\cite{li2017instance} (a), which is labeled with discontinuous coarse boundaries, and MS-COCO dataset~\cite{lin2014microsoft} (c), which is labeled with polygons, our SOC dataset (b \& d) is labeled with smooth fine boundaries.
  }\label{fig:HighQualityAnnotation}
\end{figure}
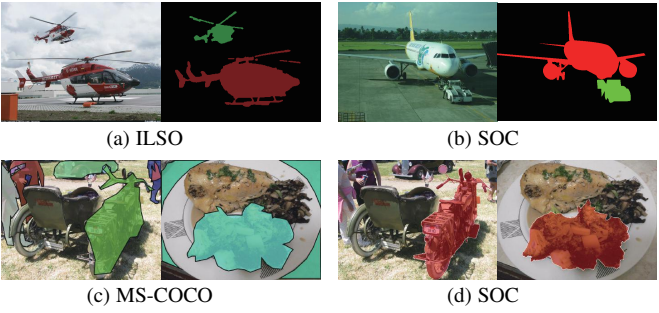

\textbf{7) High-Quality Salient Object Labeling.}
As noted in \cite{HouPami19Dss}, training on
the ECC dataset (1,000 images) yields better results than when using
other datasets (\eg, M10K with 10,000 images).
This is because, besides the scale, dataset quality is also an important.
To obtain a large number of high-quality images,
we randomly selected images from the MS-COCO dataset~\cite{lin2014microsoft},
which is a large-scale challenging dataset
whose objects are annotated with polygons (\ie, coarse labeling).
High-quality labels also play a
critical role in improving the accuracy of SOD models \cite{achanta2009frequency}.
Towards this end, we re-labeled the dataset with pixel-wise annotations.
Following other famous task-oriented SOD benchmark datasets~\cite{achanta2009frequency,alpert2007image,ChengPAMI15,jiang2017joint,li2017instance,li2015visual,LiuSZTS07Learn,2001iccvSOD,wang2017learning,xia2017and,yan2013hierarchical},
we did not use an eye tracking device.
We took two steps to ensure high-quality annotations:
(i) We asked five viewers to annotate objects that they thought were salient in each image with bounding boxes (bboxes), and
(ii) we kept the images in which the majority ($\geq 3$) of viewers
annotated the same objects (IOU of the bbox $>0.8$).
After this first stage, we had
3,000 salient object images annotated with bboxes.
\textit{In the second stage}, we further manually labeled accurate silhouettes
of the salient objects according
to the bboxes. Note that we had 10 volunteers involved in both steps to
cross-check the quality of annotations.
In the end, we kept 3,000 images with high-quality, instance-level labeled
salient objects. As shown in \figref{fig:HighQualityAnnotation} (b \& d),
the boundaries of our object labels are precise, sharp and smooth. During the
annotation process, we also added some new categories
(\eg, \emph{computer monitor, hat, pillow}) that are not labeled in
the MS-COCO dataset~\cite{lin2014microsoft}.

\section{Our Dataset-Enhancement Strategies}
Instead of focusing on designing a strong decoder for feature aggregation, we introduce three simple dataset-enhancement strategies to achieve better model generalization ability. We argue that the proposed strategies are easy to implement by existing fully supervised SOD models, and yield good performance with little effort.
Let us define the RGB saliency training dataset as $D=\{x_i,y_i\}_{i=1}^N$, where $x_i,y_i$ are an input RGB image and its corresponding ground-truth (GT) saliency map, $i$ indexes the training images, and $N$ is the size of the training dataset. As SOD is a binary prediction task, the GT saliency map $y$ is usually a binary map, and most existing SOD techniques employ a binary (or weighted) cross-entropy loss function to evaluate the saliency prediction.
In this paper, instead of defining the GT saliency map as a binary segmentation map, we first introduce \enquote{label smoothing} \cite{rethink_inception} as an effective technique to achieve both efficient model training and high model performance. Then, we adopt random image augmentation to generate diverse samples for better model generalization ability. Finally, as a widely studied technique in semi-supervised or unsupervised learning \cite{unsupervised_self_supervised_rotation, feng2019self}, we extend the self-supervised learning solution to fully supervised SOD to achieve a robust model.

\subsection{Label Smoothing}
\label{label-smo-sec}

\textbf{Label Smoothing and Knowledge Distillation.}
One of the most important scenarios in which to apply label smoothing is the teacher-student net \cite{yuan2020revisiting} for knowledge distillation.
Typically, in a teacher-student net, the teacher model has a strong learning capacity, while the student model has a lower one.
The teacher model then teaches the student model by providing the latter with a
\enquote{soft target}.
As discussed in \cite{knowledge_distilling}, the \enquote{soft target}
contains a rich similarity structure over the data,
which is essential for producing an enhanced student model.
Further, label smoothing can be treated as a form of output distribution
regularization that prevents the network from overfitting.
As pointed out in \cite{rethink_inception},
hard labels may lead to the overfitting as the model will assign full probability to each category,
which is not guaranteed to generalize well.
With soft labels, the model learns the structure of the data,
thus preventing it from being over-confident.
Following the same data setting, \eg, employing label smoothing,
\cite{Zhang2020DelvingDI} introduced online label smoothing solution
to gradually update the soft labels based on the model's prediction.

\textbf{Conventional Setting.}
Given an input image $x$ and the corresponding ground-truth saliency map $y$,
the conventional deep saliency model $f_\theta$ is trained to achieve
saliency prediction $s = f_\theta(x)$ by minimizing the cross-entropy loss:
$\mathcal{L}_{ce}(y,s)=-\sum_{i=1}^N \sum_{u,v} y_i^{u,v}\log s_i^{u,v}$,
where $(u,v)$ index pixels.
For the hard label based framework, we have $y\in\{0,1\}$,
where 1 indicates the salient foreground and 0 represents the background.

\textbf{Label Smoothing Setting.}
Different from the above hard label setting,
in label smoothing regularization (LSR) \cite{rethink_inception},
a smoothed label $y'$ is used instead of $y$, which is formulated as:
\begin{equation}\label{label_smoothing}
  y' = (1-\epsilon)y+\epsilon u(x).
\end{equation}
Here, $\epsilon$ is the smoothing parameter,
and $u(x)$ is a fixed distribution,
which is usually defined as a uniform distribution.
The smoothed label with a uniform distribution $u(x)$ is then defined as:
\begin{equation}\label{label_smoothing_uniform_ux}
    y' = (1-\epsilon)y+\frac{\epsilon}{K},
\end{equation}
where $K$ is the number of categories.

\textbf{Loss Function.}
Given smoothed label $y'$ and hard label $y$,
the loss function with LSR is defined as:
\begin{equation}\label{lsr_loss}
  \mathcal{L}_{ls} =
  (1-\alpha)\mathcal{L}_{ce}(y,s)+\alpha\mathcal{L}_{ce}(y',s),
\end{equation}
where $\alpha$ is used to balance the contribution of the smoothed and hard labels, and the smoothed label related loss is defined as
$\mathcal{L}_{lsr} = \mathcal{L}_{ce}(y',s)$.
Note that, if there exist other loss functions,
the smoothed label can only be used in cross-entropy loss.

\textbf{What Does Label Smoothing Really Do?}
The conventional cross-entropy loss can be rewritten as:
\begin{equation}\label{conventional_ce}
    \mathcal{L}_{ce} = -\log s.
\end{equation}
Here, $s$ is the model prediction after sigmoid activation (for binary classification), which is defined as:
\begin{equation}\label{pred_logits}
  s_j=e^{z_j}/\sum_{k=1}^K e^{z_k}=1/(1+\sum_{k\neq j}e^{z_k-z_j}).
\end{equation}
We then substitute $s$ in \eqnref{conventional_ce} and obtain:
\begin{equation}\label{sub_s_conv_ce}
  \mathcal{L}_{ce} = \log(1+\sum_{k\neq j}e^{z_k-z_j}).
\end{equation}
Let us define the gap between the correct class and others
as
$M = z_k-z_j$. We can then conclude that the conventional cross-entropy loss aims to maximize this gap.

For label smoothing setting, as in \eqnref{label_smoothing_uniform_ux},
we rewrite the smoothed label related loss $\mathcal{L}_{lsr}$ as:

\begin{equation}\label{label_smoothing_loss_part}
\begin{aligned}
  \mathcal{L}_{lsr}=&-((1-\epsilon)y+\epsilon/K)\log s - (1-(1-\epsilon)y-\epsilon/K)\log(1-s)\\
  =&-(y\log s +(1-y)\log (1-s))+(\epsilon y-\frac{\epsilon}{K})\log (\frac{s}{1-s}).
\end{aligned}
\end{equation}
Using the definition of $s$ in \eqnref{pred_logits}, we have:
\begin{equation}\label{second_part_lsr}
  \frac{s_j}{1-s_j} = \frac{1}{\sum_{k=1}^K e^{z_k-z_j}-1}.
\end{equation}
We can then combine \eqnref{second_part_lsr} with
\eqnref{label_smoothing_loss_part} and obtain:
\begin{equation}\label{final_lsr_loss}
  \mathcal{L}_{lsr} = \mathcal{L}_{ce}(y,s)+(\epsilon y-\frac{\epsilon}{K})*\frac{1}{\sum_{k=1}^K e^{z_k-z_j}-1}.
\end{equation}
The first part of \eqnref{final_lsr_loss} aims to maximize the gap between
the correct class and the others, which is same as the conventional binary-cross entropy loss as in \eqnref{sub_s_conv_ce}.
The second part works in the opposite direction (compared with \eqnref{sub_s_conv_ce}) to
narrow the gap. In this way, the smoothed label related loss works to balance the gap between the correct class and others, which serves as an regularization to prevent the model from being over-confident.

\subsection{Data Augmentation}
\label{data-aug-sec}

As an effective data pre-processing technique,
data augmentation aims to generate new samples from an existing dataset,
thus producing a model with good generalization ability.
Given the training dataset $D=\{x_i,y_i\}_{i=1}^N$,
data augmentation produces a new dataset
$D'=\{x'_i,y'_i\}_{i=1}^{N'}$.
As discussed previously, two main types of data augmentation have received particular attention. These include the handcrafted policies and learned policies \cite{li2020dada,AutoAugment}. For the learned policies, we observe that the augmented data can change drastically depending on the context, which may not be an issue for image classification,
but will change the salient attributes of an image.
We thus focus only on handcrafted policies.

For handcrafted data augmentation policies, existing works \cite{wei2017object,zhong2020random,chang2020mixup,guo2019mixup}
focus on three main directions:
1) image transformation, \eg, scale or rotation transformation;
2) mix-up to generate new samples, which are neighbors of the existing samples; and
3) adding noise to the ground-truth.
Similar to learned policies, the mix-up strategy change the context information of an image,
which is harmful for context-based tasks, such as salient object detection.
In this paper, we therefore focus on two very simple data augmentation techniques,
namely image transformation and adding noise to the ground-truth.
For image transformation, we randomly scale, rotate and crop part of the image
(85\% of the original image to keep the context information).
For the additive noise solution, we randomly add Gaussian noise of distribution $\mathcal{N}(0.1,0.3)$
to the ground-truth saliency map,
leading to a noisy ground-truth map.
Note that, for image transformation, we transform image and ground-truth at the
same time, while when adding noise to the ground-truth,
we only process the ground-truth saliency maps.

\subsection{Self-Supervised Learning}
\label{self-sup-learn}

Self-supervised learning learns from an image without knowing the task
itself or the ground-truth, making it an unsupervised representation learning technique.
Conventionally, for the supervised learning setting, the loss function
is defined as $\mathcal{L}_{ce}(y,s)$, where $s$ is the model's prediction,
and $y$ is the ground-truth map.
For self-supervised learning,
the final loss function usually includes two main parts:
the conventional cross-entropy loss $\mathcal{L}_{ce}(y,s)$ and
an unsupervised loss that serves as a regularizer, \ie, $\mathcal{L}(g(x),s)$,
where $g(x)$ is the transformation of the original input image $x$.
The two studies~\cite{feng2019self,zhai2019s4l}
introduced a self-supervised loss with rotation estimation as a pretext task.

Similarly, we introduce a scale/rotation consistency loss function to achieve
scale/rotation invariant predictions.
Specifically, given an input image $x$, we define its prediction as $s$.
Then, we apply an image transformation (scale or rotation transformation)
and obtain $x^t$.
We then perform the same transformation on the prediction $s$ and obtain $s'$.
We feed $x^t$ to the same salient object detection network to get the
saliency prediction as $s^t$.
We assume that $s'$ and $s^t$ should be similar.
Then, we adopt the single scale structural similarity index measure (SSIM) \cite{godard2017unsupervised,wang2004image}
as a similarity measure,
and define the self-supervised loss as:
\begin{equation}\label{self-sup-loss}
  \mathcal{L}_{ss} = 1-SSIM(s',s^t).
\end{equation}

\subsection{Loss Function with the Proposed Strategies.}
With the three introduced data-enhancement strategies,
we first apply random data augmentation to both our training image set
and training ground-truth set, as in Section \ref{data-aug-sec}.
Then we generate the smoothed label following \eqnref{label_smoothing},
with $K=2$ in this paper to represent the salient foreground and background regions.
In addition to the loss function in \eqnref{lsr_loss},
we also introduce a self-supervised loss $\mathcal{L}_{ss}$.
Our final loss function is then defined as:
\begin{equation}\label{final_loss}
  \mathcal{L} = \mathcal{L}_{ls}+\gamma \mathcal{L}_{ss},
\end{equation}
where $\gamma$ is introduced to balance the self-supervised loss,
and is empirically set to $\gamma=0.3$ in this paper.

\section{SOC Benchmark}\label{sec:benchmark}
Based on three criteria (\ie, representative pipeline, open-sourced,
and state-of-the-art performance), we select 46 traditional SOD methods
and 54 deep learning models from 203 reviewed methods
(see \secref{sec:relatedworks}) to conduct our benchmark.
To the best of our knowledge, this benchmark is the most comprehensive study
in the RGB SOD.

\subsection{Experimental Setup}
\subsubsection{Evaluation Metrics}

Note that the GTs of non-salient images in our SOC dataset are
all-zero matrices,
so directly using the traditional F-measure~\cite{achanta2009frequency}
will result in very low and inaccurate scores.
Thus, we utilize three golden metrics (\ie, MAE~\cite{fu2020siamese},
maximum E-measure~\cite{Fan2018Enhanced},
and S-measure~\cite{fan2017structure}) to avoid this issue and to provide
a more reliable assessment.
Evaluation toolboxes are now publicly available.\footnote{\url{https://github.com/mczhuge/SOCToolbox}.}
\begin{itemize}
  \item \textbf{MAE (${M}$)} is the mean absolute error metric,
    which is widely used to measure the pixel-level difference
    between the prediction and the GT.
  \item \textbf{E-measure ($E_{\xi}^{max}$)} is a new perceptual metric
    that takes both local and global similarity into consideration.
  \item \textbf{S-measure ($S_{\alpha}$)} is a standard metric that
    quantizes the structural similarity at a region and object level.
\end{itemize}

\begin{table}[thbp]
  \centering
  \footnotesize
  \renewcommand{\arraystretch}{0.5}
  \renewcommand{\tabcolsep}{1.5mm}
  \caption{SOC dataset used in the benchmarking experiments.
  }\label{tab:SOCSplit}
  \vspace{-10pt}
  \begin{tabular}{l|rrr|r}
  \toprule
    & SOC\_train & SOC\_val & SOC\_test & Total \\
  \midrule
  Salient Objects (Sal)    &  1,800 & 600 & 600 & 3,000\\
  Non-Salient Objects (NonSal) &  1,800 & 600 & 600 & 3,000 \\
  \hline
  Total               &  3,600 & 1,200 & 1,200  & 6,000\\
  \bottomrule
  \end{tabular}
\end{table}

\subsubsection{Training and Testing Protocols}
The statistics of the SOC dataset used in the benchmark are summarized
in \tabref{tab:SOCSplit}.
For traditional algorithms,
we directly test their performance on the SOC-test set (1,200 images).
For deep learning models, we first adopt the pre-trained models with
their recommended training parameter settings under the default training
dataset (see Tables \ref{tab:CNNModelSummary} \& \ref{tab:CNNModelSummary2})
and then evaluate them on the SOC\_test set to roughly obtain the 100
representative models (see \tabref{tab:TraditionalSODScore} \& \ref{tab:DeepSODScore}).
Finally, we provide a quantitative comparison and detailed analysis of
15 SOTA approaches,
including the top-5 traditional methods and top-10 deep learning models.

\begin{table}[t!]
  \centering
  \scriptsize
  \renewcommand{\arraystretch}{0.1}
  \renewcommand{\tabcolsep}{1.7mm}
  \caption{Comparison of the traditional SOD algorithms on our SOC\_test set (1,200 images) in terms of $S_{\alpha}\uparrow$, $E_{\xi}^{max}\uparrow$, and ${M}\downarrow$
  The top-3 results are highlighted in \trb{red}, \tbb{blue} and \tgb{green}, respectively.
  The superscript of each score is the corresponding ranking.
  Details of these methods are summarized in \tabref{tab:HandCraftModelSummary}.
  The overall rank index indicates the average ranking of the three metrics.
  These results are available at: \href{https://drive.google.com/drive/folders/1eubOw08o_TJyn6zYh-vHw5JpwcdhuwCg?usp=sharing}{Google Drive}.
  }\label{tab:TraditionalSODScore}
  \vspace{-10pt}
  \begin{tabular}{c|c|r|c|l|l|l|c}
  \toprule
   & \#  & Model & Code & $S_{\alpha}\uparrow$ & $E_{\xi}^{max}\uparrow$ & ${M}\downarrow$ & Rank \\
  \midrule
  \multirow{25}{*}{\vspace{-2in}\begin{sideways}2014-before\end{sideways}}
  &1&  SUN~\cite{zhang2008sun}           & \href{http://cseweb.ucsd.edu/~l6zhang/}{Matlab}     & $0.475^{46}$ & $0.688^{44}$ & $0.436^{46}$ & 46\\%-45.33 \\
  &2& LSSC~\cite{xie2012bayesian}       & \href{https://github.com/huchuanlu/13_6}{Matlab + C}%TIP
  & $0.552^{45}$ & $0.714^{43}$ & $0.365^{45}$  & 45\\%-44.33\\
  &3& BSF~\cite{sun2012saliency}        & \href{https://github.com/huchuanlu/12_13}{Matlab}%ICIP
  & $0.554^{44}$ & $0.728^{38}$ & $0.353^{44}$  & 44\\%-42\\
  &4& GR~\cite{yang2013graph}           & \href{https://github.com/huchuanlu/13_9}{Matlab + C}
  & $0.588^{41}$ & $0.715^{42}$ & $0.332^{42}$ & 43\\%-41.67\\
  &5& HS~\cite{yan2013hierarchical}     & \href{http://www.cse.cuhk.edu.hk/~leojia/projects/hsaliency/}{EXE}    & $0.601^{40}$ & $0.729^{37}$ & $0.321^{41}$ & 42\\%-39.3\\
  &6&  Itti~\cite{itti1998model}         & \href{http://ilab.usc.edu/toolkit/}{Matlab}   & $0.587^{42}$ & $0.736^{30}$ & $0.311^{39}$ & 41\\%-37 \\
  &7&  AIM~\cite{bruce2006saliency}      & \href{https://github.com/TsotsosLab/AIM}{Matlab}
  & $0.605^{39}$ & $0.670^{45}$ & $0.250^{24}$ & 39\\%-36\\
  &8&  GBVS~\cite{harel2007graph}        & \href{http://people.vision.caltech.edu/~harel/share/gbvs.php}{Matlab}%NeurIPS
  & $0.615^{36}$ & $0.733^{35}$ & $0.293^{37}$ & 39\\%-36.33\\
  &9& LR~\cite{shen2012unified}         & \href{https://xiaohuishen.github.io/assets/code_lowranksaliency.zip}{Matlab}%CVPR
  & $0.642^{31}$ & $0.723^{40}$ & $0.253^{27}$ & 36\\%-33-32.6\\ 27-28
  &10&  CA~\cite{goferman2011context}     & \href{https://github.com/MCG-NKU/SalBenchmark/tree/master/Code/matlab/CA}{Matlab + C}    & $0.606^{38}$ & $0.750^{22}$ & $0.291^{36}$ & 35\\%-31.3 \\
  &11& MR~\cite{YangZLRY13Manifold}      & \href{https://github.com/huchuanlu/13_4}{Matlab + C}%CVPR
  & $0.645^{29}$ & $0.734^{33}$ & $0.259^{31}$ & 32\\%-31.33 \\
  &12&  SEG~\cite{rahtu2010segmenting}    & \href{https://github.com/MCG-NKU/SalBenchmark/tree/master/Code/matlab/SEG}{Matlab + C}    & $0.576^{43}$ & $0.765^7$ & $0.352^{43}$ & 32\\%-31 \\
  &13&  FT~\cite{achanta2009frequency}    & \href{https://ivrlwww.epfl.ch/supplementary_material/RK_CVPR09/}{C}    & $0.626^{34}$ & $0.738^{29}$ & $0.236^{20}$ & 28\\%-27.67 \\
  &14& MC~\cite{jiang2013saliency}       & \href{https://github.com/huchuanlu/13_2}{Matlab + C}    & $0.656^{23}$ & $0.736^{30}$ & $0.251^{25}$ & 26\\%-26.33\\ 25-26
  &15& CB~\cite{jiang2011automatic}      & \href{http://jianghz.me/files/CBSaliency-release.zip}{Matlab + C}    & $0.653^{25}$ & $0.758^{13}$ & $0.268^{33}$ & 23\\%-23.67,26.33\\
  &16&  SR~\cite{hou2007saliency}         & \href{https://github.com/uoip/SpectralResidualSaliency}{Matlab/C++}
  & $0.658^{21}$ & $0.661^{46}$ & $0.156^4$ & 23\\%-23.67\\
  &17& PCA~\cite{margolin2013makes}      & \href{https://github.com/MCG-NKU/SalBenchmark/tree/master/Code/matlab/PCA}{Matlab + C}    & $0.670^{18}$ & $0.741^{28}$ & $0.209^{13}$ & 17\\%-20    \\
  &18& MSS~\cite{tong2014saliency}       & \href{https://github.com/huchuanlu/14_10}{Matlab}%SPL
  & $0.682^{12}$ & $0.776^4$ & $0.231^{19}$ & 10\\%-11.67\\
  &19& SF~\cite{perazzi2012saliency}     & \href{https://fperazzi.github.io/projects/saliency_filters/files/saliencyfilters.zip}{C}    & $0.699^6$ & $0.747^{26}$ & $\trb{0.130}^1$ & 8\\%-11\\
  &20& DSR~\cite{li2013saliency}         & \href{https://github.com/huchuanlu/13_1}{Matlab + C}    & $0.702^5$ & $0.751^{20}$ & $0.184^8$ & 8\\%-11.67\\
  &21&  MSSS~\cite{achanta2010saliency}  & \href{https://www.epfl.ch/labs/ivrl/research/saliency/saliency-msss/}{C}    & $0.683^{11}$ & $0.757^{14}$ & $0.164^5$ & 7\\%-10 \\
  &22& HDCT~\cite{kim2014salient}        & \href{https://sites.google.com/site/kjw02040/hdct}{Matlab}%CVPR
  & $0.696^7$ & $0.774^5$ & $0.201^{12}$ & 6\\%-8.33\\
  &23& DRFI~\cite{jiang2013salient}      &  \href{http://jianghz.me/drfi/}{C}   & $0.709^4$ & $\tbb{0.791}^2$ & $0.197^{11}$ & 4\\%-5.67\\
  &24& COV~\cite{erdem2013visual}        & \href{https://web.cs.hacettepe.edu.tr/~erkut/projects/CovSal/}{Matlab}     & $\tgb{0.711}^3$ & $0.761^{9}$ & $\tbb{0.146}^2$ & 2\\%-4.67\\
  &25& RBD~\cite{zhu2014saliency}        & \href{https://github.com/MCG-NKU/SalBenchmark/tree/master/Code/matlab/RBD}{Matlab}    & $\tbb{0.716}^2$ & $\tgb{0.784}^3$ & $0.186^{9}$ & 2\\%-4.67\\
  \midrule
  \multirow{21}{*}{\vspace{-1.6in}\begin{sideways}2021-2015\end{sideways}}
  &26& WMR~\cite{zhu2018saliency}        & \href{https://pan.baidu.com/s/1NFboaeuuBph_QQGgXuIcaA}{Matlab + C}
  & $0.640^{32}$ & $0.733^{35}$ & $0.269^{34}$ & 38\\%-34.33\\
  &27& MAPM~\cite{sun2015saliency}       & \href{https://github.com/huchuanlu/15_6}{Matlab + C}%TIP
  & $0.644^{30}$ & $0.722^{41}$ & $0.256^{29}$ & 37\\%-33.67,33.667\\ 29-30
  &28& BL~\cite{tong2015salient}         & \href{https://github.com/huchuanlu/15_13}{Matlab + C}%CVPR
  & $0.623^{35}$ & $0.751^{20}$ & $0.296^{38}$ & 32\\%-31.67\\
  &29& RRWR~\cite{li2015robust}          & \href{https://github.com/yuanyc06/rr}{Matlab}%CVPR
  & $0.647^{27}$ & $0.735^{32}$ & $0.258^{30}$ & 31\\%-27.67\\ 30-24
  &30& WLRR~\cite{tang2016salient}       & \href{http://tangchang.net/codes/WLRRSalDemo.zip}{Matlab + C}%SPL
  & $0.614^{37}$ & $0.759^{11}$ & $0.312^{40}$ & 30\\%-29.33\\
  &31& RCRR~\cite{yuan2017reversion}     & \href{https://github.com/yuanyc06/rcrr}{Matlab}%TIP
  & $0.650^{26}$ & $0.734^{33}$ & $0.255^{28}$ & 29\\%-29.67\\ 28-29
  &32& GP~\cite{jiang2015generic}        & \href{http://www.svcl.ucsd.edu/publications/}{Matlab + C}%ICCV
  & $0.632^{33}$ & $0.759^{11}$ & $0.287^{35}$ & 27\\%-26.33\\
  &33& TLLT~\cite{gong2015saliency}      & \href{http://www.kerenfu.top/}{Matlab} %CVPR
  & $0.656^{23}$ & $0.725^{39}$ & $0.214^{15}$ & 25\\%-25.67\\
  &34& BSCA~\cite{qin2015saliency}       & \href{http://cseweb.ucsd.edu/~yaq007/code-CA15.zip}{Matlab + C}%CVPR
  & $0.657^{22}$ & $0.755^{16}$ & $0.259^{31}$ & 22\\%-23\\
  &35& SMD~\cite{peng2016salient}        & \href{https://sites.google.com/site/salientobjectdetection/need-to-knows}{Matlab}%TPAMI
  & $0.662^{20}$ & $0.748^{25}$ & $0.246^{22}$ & 21\\%-22.33\\
  &36& MDC~\cite{huang2017300}           & \href{https://github.com/huangxm14-thu/SaliencyMDC}{C}%TIP
  & $0.675^{16}$ & $0.744^{27}$ & $0.219^{17}$ & 20\\%-20\\
  &37& DSP~\cite{chen2016discriminative} & \href{https://github.com/ShuhanChen/DSP_PR2016}{Matlab + C}%PR
  & $0.664^{19}$ &  $0.754^{17}$ & $0.248^{23}$ & 17\\%-19.67\\
  &38& MIL~\cite{huang2017salient}       & \href{https://github.com/huchuanlu/17_8}{Matlab + C}%TIP
  & $0.671^{17}$ & $0.750^{22}$ & $0.236^{20}$ & 17\\%-19\\
  &39& MST~\cite{tu2016real}             & \href{https://github.com/lhaof/Real-Time-Salient-Object-Detection-with-a-Minimum-Spanning-Tree}{C}
  & $0.647^{27}$ & $0.773^6$ & $0.251^{25}$ & 16\\%-16.67,19.33\\ 25-26
  &40& GLC~\cite{tong2015salientPR}      & \href{https://github.com/huchuanlu/15_13}{Matlab + C}%PR
  & $0.676^{15}$ & $0.756^{15}$ & $0.223^{18}$ & 15\\%-16\\
  &41& MBS~\cite{zhang2015minimum}       & \href{https://github.com/jimmie33/MBS}{Matlab}%ICCV
  & $0.678^{14}$ & $0.753^{18}$ & $0.214^{15}$ & 14\\%-11.67\\
  &42& LPS~\cite{li2015inner}            & \href{https://github.com/huchuanlu/15_7}{Matlab + C}%TIP
  & $0.694^{9}$ & $0.749^{24}$ & $0.183^7$ & 13\\%-13.33\\
  &43& WFD~\cite{huang2018water}         & \href{https://github.com/huangxm14-thu/SaliencyWater}{C}%PR
  & $0.680^{13}$ & $0.760^{10}$ & $0.213^{14}$ & 12\\%-11.67 \\
  &44& BFS~\cite{wang2015saliencyNC}     & \href{https://github.com/huchuanlu/15_16}{Matlab + C}%NC
  & $0.696^7$ & $0.753^{18}$ & $0.195^{10}$ & 10\\%-11.67\\
  &45& WSC~\cite{li2015weighted}         & \href{https://github.com/aistairc/SC_based_gaze_prediction}{Matlab}%CVPR
  & $0.693^{10}$ & $0.765^7$ & $0.179^6$ & 5\\%-7.67 \\
  &46& HCCH~\cite{liu2017hierarchical}   & \href{https://sites.google.com/view/hcchsal}{Matlab}%TIP
  & $\trb{0.736}^1$ & $\trb{0.794}^1$ & $\tgb{0.149}^3$ & 1\\%-1.67\\
  \bottomrule
  \end{tabular}
\end{table}

\begin{table}[t!]
  \centering
  \scriptsize
  \renewcommand{\arraystretch}{0.1}
  \renewcommand{\tabcolsep}{1.5mm}
  \caption{
  Evaluation of 54 deep learning based SOD models on our SOC\_test set (1,200 images).
  We adopt the default implementations listed in \tabref{tab:CNNModelSummary} and \tabref{tab:CNNModelSummary2} to test their generalization capability.
  These results are available at: \href{https://drive.google.com/drive/folders/1AF1w7auAE3y1qT3FuYr_wae20okB7BUm}{Google Drive}.
  }\label{tab:DeepSODScore}
  \vspace{-10pt}
  \begin{tabular}{c|c|r|c|l|l|l|c}
  \toprule
   & \#  & Model & Code & $S_{\alpha}\uparrow$ & $E_{\xi}^{max}\uparrow$ & ${M}\downarrow$ & Rank \\
  \midrule
  \multirow{3}{*}{\begin{sideways}2015\end{sideways}\vspace{-.2in}}
  &1&  LEGS~\cite{wang2015deep}      & \href{https://github.com/huchuanlu/15_3}{Caffe}%CVPR
  & $0.679^{53}$ & $0.765^{54}$ & $0.228^{53}$ & 54\\%-53\\
  &2&  MDF~\cite{li2015visual}       & \href{https://sites.google.com/site/ligb86/mdfsaliency/}{Caffe}%CVPR
  & $0.739^{49}$ & $0.768^{53}$ & $0.144^{43}$ %error: $0.271^{53}$
  & 49\\%-51.66\\
  &3&  MC~\cite{zhao2015saliency}    & \href{https://github.com/Robert0812/deepsaldet}{Caffe}%CVPR
  & $0.757^{47}$ & $0.823^{43}$ & $0.138^{35}$ & 43\\%-41.66\\
  \midrule
  \multirow{7}{*}{\vspace{-.5in}\begin{sideways}2016\end{sideways}}
  &4&  DSL~\cite{yuan2016dense}      & \href{https://github.com/yuanyc06/dsl}{Caffe}%TCSVT
  & $0.724^{52}$ & $0.810^{47}$ & $0.194^{52}$ & 51\\%-50\\
  &5&  DISC~\cite{chen2016disc}      & \href{https://github.com/liulingbo918/DISC}{Caffe}%TNNLS
  & $0.735^{51}$ & $0.810^{47}$ & $0.175^{50}$ & 50\\%-49\\
  &6& DCL~\cite{li2016deep}         & \href{https://sites.google.com/site/ligb86/hkuis}{Caffe}%CVPR
  & $0.771^{44}$ & $0.836^{39}$ & $0.157^{48}$ & 45\\%-43.33\\
  &7& ELD~\cite{lee2016deep}        & \href{https://github.com/gylee1103/SaliencyELD}{Caffe}%CVPR
  & $0.774^{42}$ & $0.836^{39}$ & $0.138^{35}$ & 40\\%-38.66\\
  &8&  DS~\cite{li2016deepsaliency}  & \href{https://github.com/zlmzju/DeepSaliency}{Caffe}%TIP
  & $0.779^{40}$ & $0.860^{24}$ & $0.155^{46}$ & 37\\% -36.33\\
  &9& DHS~\cite{liu2016dhsnet}      & \href{https://github.com/wlguan/DHSNet-PyTorch}{Pytorch}%CVPR
  & $0.800^{32}$ & $0.848^{33}$ & $0.122^{30}$ & 33\\%-31.66\\
  &10& RFCN~\cite{wang2016saliency}  & \href{https://github.com/huchuanlu/16_1}{Caffe}%ECCV
  & $0.814^{23}$ & $0.858^{27}$ & $0.113^{23}$ & 25\\%-24.33\\
  \midrule
  \multirow{8}{*}{\vspace{-.5in}\begin{sideways}2017\end{sideways}}
  &11& UCF~\cite{zhang2017learning}  & \href{https://github.com/Pchank/caffe-sal}{Caffe}%ICCV
  & $0.654^{54}$ & $0.805^{51}$ & $0.285^{54}$  & 53\\%-53\\
  &12& AMU~\cite{zhang2017amulet}    & \href{https://github.com/Pchank/caffe-sal}{Caffe}%ICCV
  & $0.737^{50}$ & $0.808^{50}$ & $0.185^{51}$ & 51\\%-50 \\
  &13& SVF~\cite{zhang2017supervis}  & \href{https://github.com/zhangyuygss/SVFSal.caffe}{Caffe}%ICCV
  & $0.761^{45}$ & $0.816^{45}$  & $0.156^{47}$ & 47\\%-45.33\\
  &14& WSS~\cite{wang2017learning}   & \href{https://github.com/scott89/WSS}{Caffe}%CVPR
  & $0.778^{41}$ & $0.821^{44}$  & $0.140^{39}$ & 42\\%-41.33\\
  &15& DSS~\cite{hou2017dss}         & \href{https://github.com/Andrew-Qibin/DSS}{Caffe}%CVPR
  & $0.807^{30}$ & $0.858^{27}$  & $0.111^{20}$ & 27\\%-25.66\\
  &16& SRM~\cite{wang2017stagewise}  & \href{https://github.com/Pchank/caffe-sal}{Caffe}%CVPR
  & $0.822^{16}$ & $0.859^{26}$  & $0.111^{20}$ & 21\\%-20.66\\
  &17& MSRNet~\cite{li2017instance}  & \href{https://github.com/Xyuan13/MSRNet}{Caffe}%CVPR
  & $0.816^{19}$ & $0.871^{16}$  & $0.117^{25}$ & 20\\%-20 \\
  &18& NLDF~\cite{Luo2017CVPR}       & \href{https://github.com/zhimingluo/NLDF}{Tensorflow}%CVPR
  & $0.816^{19}$ & $0.860^{24}$  & $0.104^{13}$ & 16\\%-18.66\\
  %&25& FSN~\cite{chen2017look}       & ICCV  &       &         \\
  \midrule
  \multirow{8}{*}{\vspace{-.5in}\begin{sideways}2018\end{sideways}}
  &19& RAS~\cite{chen2018reverse}      & \href{https://github.com/ShuhanChen/RAS_ECCV18}{Pytorch}%ECCV
  & $0.759^{46}$ & $0.813^{46}$ & $0.151^{44}$ & 46\\%-45\\
  &20& R3Net~\cite{deng2018r3net}      & \href{https://github.com/zijundeng/R3Net}{Pytorch}%IJCAI
  & $0.773^{43}$ & $0.825^{42}$ & $0.138^{35}$ & 41\\%-40\\

  &21& LPSNet~\cite{zeng2018learning}  & \href{https://github.com/zengxianyu/lps}{Pytorch}%CVPR
  & $0.795^{35}$ & $0.838^{38}$ & $0.143^{42}$ & 39\\%-38.33\\
  &22& DGRL-GLN~\cite{wang2018detect}  & \href{https://github.com/TiantianWang/CVPR18_detect_globally_refine_locally}{Caffe}%CVPR
  & $0.794^{36}$ & $0.845^{36}$ & $0.141^{40}$ & 38\\%-37.33 \\
  &23& C2SNet~\cite{li2018contour}    & %C2SNet-30k
  \href{https://github.com/lixin666/C2SNet}{Caffe} %ECCV
  & $0.791^{37}$ & $0.845^{36}$ & $0.138^{35}$ & 36\\%-36\\
  &24& PiCA-Res~\cite{liu2018picanet}& \href{https://github.com/Ugness/PiCANet-Implementation}{Pytorch}%CVPR
  & $0.810^{28}$ & $0.858^{27}$ & $0.128^{31}$ & 31\\%-28.66\\
  &25& BMPM~\cite{zhang2018bi}        &
  \href{https://github.com/zhangludl/A-bi-directional-message-passing-model-for-salient-object-detection}{Tensorflow}%CVPR
  & $0.810^{28}$ & $0.853^{30}$ & $0.119^{27}$ & 29\\%-28.33\\
  &26& ASNet~\cite{wang2018salient}    & \href{https://github.com/wenguanwang/ASNet}{Keras}%CVPR
  & $0.817^{18}$ & $0.865^{20}$ & $0.111^{20}$ & 17\\%-19.33\\
  \midrule
  \multirow{9}{*}{\vspace{-.6in}\begin{sideways}2019\end{sideways}}
  &27& MWS~\cite{zeng2019multi}      & \href{https://github.com/zengxianyu/mws}{Pytorch}%CVPR
  & $0.757^{47}$ & $0.828^{41}$ & $0.172^{49}$ & 47\\%-45.33\\
  &28& AFNet~\cite{feng2019attentive} & \href{https://github.com/ArcherFMY/AFNet}{Caffe}%CVPR
  & $0.812^{24}$ & $0.850^{32}$ & $0.120^{29}$ & 29\\%-28.33 \\
  &29& SIBA~\cite{su2019selectivity} &  \href{http://cvteam.net/projects/ICCV19-SOD/BANet.html}{Caffe}  & $0.800^{32}$ & $0.884^{10}$ & $0.130^{33}$ & 26\\%-25\\
  &30& Deepside~\cite{fu2019deepside} & \href{https://github.com/kerenfu/Deepside}{Caffe} & %NC &
  $0.815^{21}$ & $0.861^{23}$ & $0.119^{27}$ & 24\\%-23.66\\
  &31& PFANet~\cite{zhao2019pyramid} & \href{https://github.com/CaitinZhao/cvpr2019_Pyramid-Feature-Attention-Network-for-Saliency-detection}{Tensorflow}%CVPR
  & $0.815^{21}$ & $0.846^{35}$ & $0.101^8$ & 22\\%-21.33\\
  &32& PoolNet~\cite{liu2019simple}  & \href{https://github.com/backseason/PoolNet}{Pytorch}%CVPR
  & $0.829^{13}$ & $0.868^{18}$ & $0.106^{16}$ & 14\\%-15.66\\
  &33& SCRNet~\cite{wu2019stacked}   & \href{https://github.com/wuzhe71/SCRN}{Pytorch}%ICCV
  & $0.833^{11}$ & $0.872^{15}$ & $0.105^{14}$ & 13\\%-13\\
  &34& CPDVgg~\cite{wu2019cascaded} & \href{https://github.com/wuzhe71/CPD}{Pytorch}%CVPR
  & $\tgb{0.856}^3$ & $0.889^6$ & $\tbb{0.079}^2$ & 2\\%-3.66\\
  &35& EGNet~\cite{ zhao2019egnet}      & \href{https://github.com/JXingZhao/EGNet}{Pytorch}%ICCV
  & $\trb{0.858}^1$ & $\tbb{0.896}^2$ & $\trb{0.078}^1$ & 1\\%-1.33 \\
  \midrule
  \multirow{14}{*}{\vspace{-1.3in}\begin{sideways}2020\end{sideways}}
  &36& ABPNet~\cite{zhang2020learning} & \href{https://github.com/JingZhang617/Noise-aware-ABP-Saliency}{Pytorch}%ECCV
  &  $0.783^{38}$ &  $0.810^{47}$ & $0.153^{45}$ & 44\\%-43\\
  &37& U2Net~\cite{qin2020u2} & \href{https://github.com/NathanUA/U-2-Net}{Pytorch}%PR
  &  $0.780^{39}$ & $0.795^{52}$ & $0.105^{14}$ & 35\\%-35\\
  &38& GCPANet~\cite{chen2020global} & \href{https://github.com/JosephChenHub/GCPANet}{Pytorch}%AAAI
  & $0.807^{30}$ & $0.848^{33}$ & $0.133^{34}$ & 34\\%-32.33\\
  &39& ITSD~\cite{zhou2020interactive} & \href{https://github.com/moothes/ITSD-pytorch}{Pytorch}%CVPR
  & $0.798^{34}$ & $0.870^{17}$ & $0.142^{41}$ & 32\\%-30.66\\
  &40& MINet~\cite{pang2020multi} & \href{https://github.com/lartpang/MINet}{Pytorch}%CVPR
  & $0.819^{17}$ & $0.864^{22}$ & $0.117^{25}$ & 22\\%-14.66\\
  &41& SANet~\cite{zhang2020weakly} & \href{https://github.com/JingZhang617/Scribble_Saliency}{Pytorch}%CVPR
  &  $0.812^{24}$ & $0.868^{18}$ & $0.106^{16}$ & 17\\%-19.33\\
  &42& GateNetVgg~\cite{zhao2020suppress} & \href{https://github.com/Xiaoqi-Zhao-DLUT/GateNet-RGB-Saliency}{Pytorch} & $0.827^{15}$ & $0.865^{20}$ & $0.108^{18}$ & 15\\%-17.66\\
  &43& F3Net~\cite{wei2020f3net} & \href{https://github.com/weijun88/F3Net}{Pytorch}%AAAI
  & $0.828^{14}$ & $0.891^5$ & $0.109^{19}$ & 12\\%-12.66\\
  &44& CSNet~\cite{gao2020highly} & \href{https://github.com/MCG-NKU/SOD100K/tree/master/CSNet}{Pytorch}%ECCV
  &  $0.834^{10}$ &  $0.876^{14}$ & $0.103^{10}$ & 11\\%-11.66\\
  &45& LDF~\cite{wei2020label} & \href{https://github.com/weijun88/LDF}{Pytorch}%CVPR
  & $0.835^9$ & $0.878^{12}$ & $0.103^{10}$ & 10\\%-10.33\\
  &46& RASNet~\cite{chen2020reverse} & \href{https://github.com/ShuhanChen/RAS-pytorch}{Pytorch}%TIP
  & $0.832^{12}$ & $0.887^8$ & $0.103^{10}$ & 9\\%-10\\
  &47& CAGVgg~\cite{mohammadi2020cagnet} & %PR &
  \href{https://github.com/Mehrdad-Noori/CAGNet}{Keras} & $0.837^8$ & $0.878^{12}$ & $0.088^4$ & 8\\%-8\\
  &48& DFI~\cite{liu2020dynamic} & \href{https://github.com/backseason/DFI}{Pytorch}%TIP
  & $0.838^7$ & $\trb{0.903}^1$ & $0.101^8$ & 5\\%-5.33\\
  &49& R2Net~\cite{feng2020residual} & %TIP &
  \href{https://github.com/ArcherFMY/R2Net}{Pytorch} & $\tbb{0.857}^2$ & $0.885^9$ & $\tgb{0.084}^3$ & 4\\
  \midrule
  \multirow{5}{*}{\vspace{-.2in}\begin{sideways}2021\end{sideways}}
  &50& SCWS~\cite{yu2021structure}  & \href{https://github.com/siyueyu/SCWSSOD}{Pytorch}%AAAI
  & $0.811^{26}$ & $0.851^{31}$ & $0.115^{24}$ & 28\\% -27\\
  &51& ICON~\cite{zhuge2021salient}    & \href{https://github.com/mczhuge/ICON}{Pytorch}
  & $0.811^{26}$ & $\tbb{0.896}^2$ & $0.128^{31}$ & 19\\%-19.66\\
  &52& BAS~\cite{qin2021boundary}      & \href{https://github.com/NathanUA/BASNet}{Pytorch}
  & $0.842^5$ & $0.882^{11}$ & $0.092^7$ & 7\\%-7.66\\
  &53& ABP~\cite{zhang2020uncertainty} & \href{https://github.com/JingZhang617/UCNet}{Pytorch} & $0.842^5$ & $0.889^6$ & $0.091^6$ & 6\\%-5.66\\
  &54& CVAE~\cite{zhang2020uncertainty} & \href{https://github.com/JingZhang617/UCNet}{Pytorch} & $0.849^4$ & $\tgb{0.892}^4$ & $0.089^5$ & 3\\%-4.33\\
  \bottomrule
  \end{tabular}
\end{table}

\subsection{Quantitative Comparisons}\label{sec:Quantitative}

To build a standardized leaderboard
(\ie, same image resolution, thresholding step, and evaluation tool),
we provide three golden metrics, \ie,
$S_{\alpha}$, $E_{\xi}^{max}$, and ${M}$.

\tabref{tab:TraditionalSODScore} shows the performance of 46 SOTA
traditional SOD  algorithms on our SOC\_test set.
In terms of both \textbf{S-measure} (\ie, $S_{\alpha}$) and
max \textbf{E-measure} ($E_{\xi}^{max}$),
the HCCH method surpasses all competitors by a large margin.
RBD, COV, and DRFI obtain comparable performance in terms of
$S_{\alpha}$ score.
Meanwhile, COV ranks third in terms of $S_{\alpha}$ measure,
but ninth in $E_{\xi}^{max}$.
In terms of \textbf{MAE} (\ie, ${M}$), the top-5 approaches are:
SF, COV, HCCH, SR, and MSSS.
It is worth mentioning that SF reduces ${M}$ and outperforms
all the recent traditional SOD methods.
Based on their overall scores, the top-5 methods are HCCH, RBD, COV, DRFI,
and WSC.

The quantitative results of the 54 deep learning SOD models on our
SOC\_test dataset are shown in \tabref{tab:DeepSODScore}.
In terms of $S_{\alpha}$, EGNet, R2Net, and CPDVgg are the top-3 models,
with scores of more than 0.85.
Roughly 46\% (\ie, 21/45) of model scores are between 0.650 and 0.800.
Compared with the traditional model, which achieves an $S_{\alpha}$ score
of 0.736, we can see continuous improvement over the past few years,
with the exception of four early models (\ie, DISC, DSL, LEGS, and UCF).
At the same time, 30 out of 45 models achieve high performance
(\eg, 0.800$\leq S_{\alpha}\leq$0.850) and the average performance is
nearly 0.820.
Interestingly, in terms of $E_{\xi}^{max}$, the multi-task learning
framework DFI and integrity learning model have the best and second-best scores of 0.903
and 0.896, respectively.
Consistent with S-measure, in terms of \textbf{MAE},
we obtain the same top-3 models EGNet, CPDVgg, and R2Net.
From our 54 benchmarked models, we find that models that perform well
in terms of S-measure also do well in MAE.
Overall, the top-10 approaches are EGNet, CPDVgg, CVAE, R2Net, DFI, ABP,
BAS, CAGVgg, RASNet, and LDF.
In the following section (\secref{sec:discussion}),
we will provide a more detailed analysis of these models.

\begin{figure}[t!]
  \centering
  \begin{overpic}[width=.8\columnwidth]{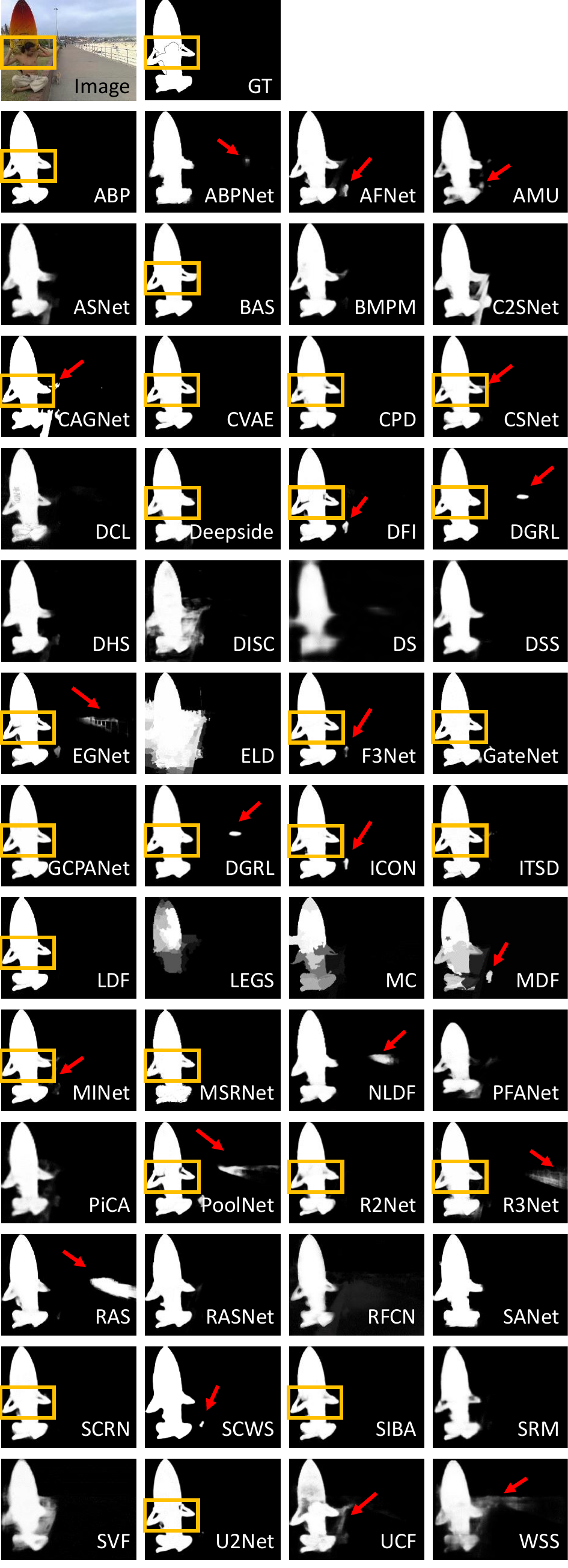}
  \end{overpic}
  \vspace{-10pt}
  \caption{
  Visualization results of deep learning models.
  }\label{fig:DeepTop50}
\end{figure}

\begin{figure}[t!]
  \centering
  \begin{overpic}[width=.8\columnwidth]{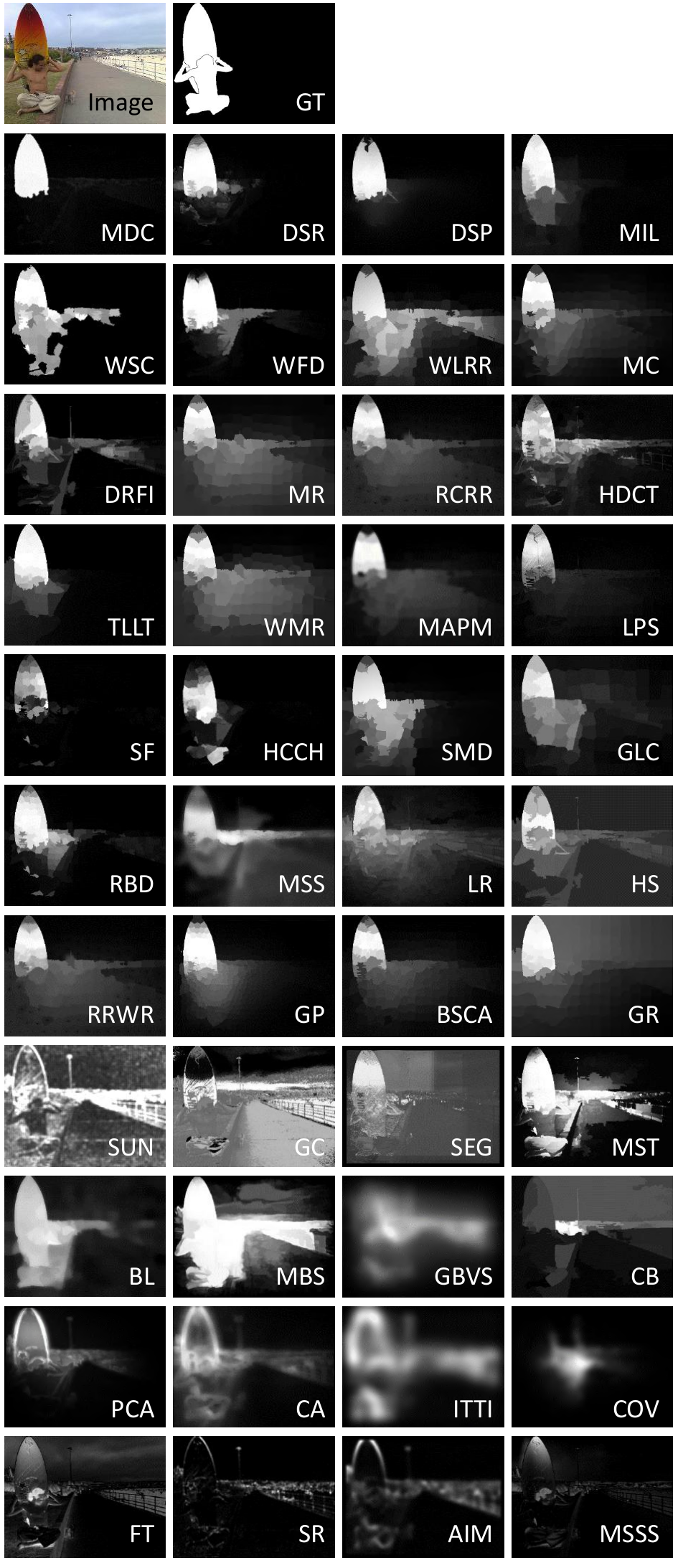}
  \end{overpic}
  \vspace{-10pt}
  \caption{Qualitative results of state-of-the-art traditional approaches.
  }\label{fig:TraditionalTop50}
  \vspace{-5pt}
\end{figure}

\subsection{Qualitative Comparisons}\label{sec:Qualitative}
Two qualitative comparisons are presented in \figref{fig:DeepTop50}
and \figref{fig:TraditionalTop50}.
As can be seen from \figref{fig:DeepTop50}, deep models generate saliency maps
that are similar to the GTs, to varying degrees.
Specifically, for ASNet, C2SNet, BMPM, DCL, DHS, DSS, DS, DISC, SVF, RFCN,
and PFANet, the position of the object can be well-identified.
However, all these methods generate blurred responses on object boundaries.
PFANet, MDF, MC and LEGS even nearly destroy the integrity of the object.
To better highlight these results, we introduce yellow rectangles to
mark the high-quality segmentation regions and utilize red arrows to point
out the errors.
We observe that eight models (ABPNet, AFNet, AMU, NLDF, RAS, SCWS, UCF,
and WSS) can localize the human object but introduce additional noise.
We also notice that CAGNet, CSNet, MINet, DGRL, EGNet, F3Net, ICON,
PoolNet, and R3Net can even capture the small structure of the human elbow.
Moreover, saliency maps from R2Net, Deepside, SIBA, and MSRNet demonstrate
better results than the above-mentioned methods.
Amazingly, BAS, U2Net, ABP, CPD, GateNet, GCPANet, ITSD, LDF, SCRN,
and CAVE perform very close to the GT and result in knife-edge-shaped
boundaries in the yellow rectangle region without any additional noise.

\begin{table*}[t!]
  \centering
  \scriptsize
  \renewcommand{\arraystretch}{0.6}
  \renewcommand{\tabcolsep}{1.13mm}
  \caption{
  Comparison of 14 state-of-the-art approaches in terms of attribute-level performance.
  For deep learning models, we re-train them on our SOC-Sal\_train set
  (\ie, 1,800 images).
  Please refer to Tables \ref{tab:HandCraftModelSummary},
  \ref{tab:CNNModelSummary}, \& \ref{tab:CNNModelSummary2} for more details.
  These results are available at: \href{https://drive.google.com/drive/folders/1c4CgQICbRKg0Hzj4XhBRHr0o9OKu2MSf?usp=sharing}{Google Drive}.
  }\label{tab:AttributeBenchmark}
  \vspace{-10pt}
  \begin{tabular}{c|r|cc|cc|cc|cc|cc|cc|cc|cc|cc|cc}
  \toprule
  & &\multicolumn{2}{c|}{AC}&\multicolumn{2}{c|}{BO}&\multicolumn{2}{c|}{CL}&\multicolumn{2}{c|}{HO}&\multicolumn{2}{c|}{MB}&\multicolumn{2}{c|}{OC}&\multicolumn{2}{c|}{OV}&\multicolumn{2}{c|}{SC}&\multicolumn{2}{c|}{SO} & \multicolumn{2}{c}{Avg.} \\
    & \diagbox[width=7em,height=2em,trim=r]{Model}{Attribute}
    & $S_{\alpha}\uparrow$&${M}\downarrow$
    & $S_{\alpha}\uparrow$&${M}\downarrow$
    & $S_{\alpha}\uparrow$&${M}\downarrow$
    & $S_{\alpha}\uparrow$&${M}\downarrow$
    & $S_{\alpha}\uparrow$&${M}\downarrow$
    & $S_{\alpha}\uparrow$&${M}\downarrow$
    & $S_{\alpha}\uparrow$&${M}\downarrow$
    & $S_{\alpha}\uparrow$&${M}\downarrow$
    & $S_{\alpha}\uparrow$&${M}\downarrow$
    & $S_{\alpha}\uparrow$&${M}\downarrow$\\
    \hline
    \multirow{5}{*}{\begin{sideways}Traditional\end{sideways}\vspace{-.1in}}
    & COV~\cite{erdem2013visual}  & 0.505 & 0.216 & 0.277 & 0.577 & 0.453 & 0.280 & 0.508 & 0.229 & 0.494 & 0.219 & 0.484 & 0.246 & 0.423 & 0.314 & 0.535 & 0.174 & 0.525 & 0.172 & 0.467 & 0.270  \\
    & WSC~\cite{li2015weighted}   & 0.541 & 0.205 & 0.356 & 0.517 & 0.517 & 0.252 & 0.556 & 0.211 & 0.536 & 0.210 & 0.529 & 0.227 & 0.475 & 0.292 & 0.567 & 0.170 & 0.535 & 0.181  & 0.512 & 0.252\\
    & HCCH~\cite{liu2017hierarchical}  & 0.585 & 0.199 & 0.354 & .0.525 & 0.537 & 0.254 & 0.615 & 0.197 & 0.547 & 0.202 & 0.552 & 0.225 & 0.468 & 0.298 & 0.595 & 0.165 & 0.588 & 0.162 & 0.538 & 0.247\\
    & DRFI~\cite{jiang2013salient}  & 0.598 & 0.229 & 0.391 & 0.513 & 0.570 & 0.274 & 0.618 & 0.230 & 0.556 & 0.230 & 0.577 & 0.248 & 0.527 & 0.304 & 0.614 & 0.188 & 0.585 & 0.197 & 0.560 & 0.268\\
    & RBD~\cite{zhu2014saliency}  & 0.589 & 0.225 & 0.429 & 0.481 & 0.575 & 0.260 & 0.625 & 0.216 & 0.557 & 0.213 & 0.583 & 0.235 & 0.521 & 0.295 & 0.602 & 0.191 & 0.579 & 0.192  & 0.562 & 0.256\\
    \midrule
    \multirow{10}{*}{\begin{sideways}Deep Learning \end{sideways}}
    & ABP~\cite{zhang2020uncertainty}  & 0.767 & 0.092 & 0.592 & 0.315 & 0.742 & 0.125 & 0.787 & 0.101 & 0.742 & 0.095 & 0.740 & 0.112 & 0.746 & 0.132 & 0.759 & 0.083 & 0.741 & 0.080 & 0.735 & 0.126 \\
    & EGNet~\cite{zhao2019egnet}  & 0.791 & 0.088 & 0.593 & 0.307 & 0.739 & 0.137 & 0.788 & 0.110 & 0.763 & 0.115 & 0.743 & 0.120 & 0.750 & 0.138 & 0.800 & 0.076 & 0.753 & 0.088  & 0.747 & 0.131\\
    & CPDVgg~\cite{wu2019cascaded} & 0.806 & 0.076 & 0.626 & 0.278 & 0.765 & 0.118 & 0.808 & 0.096 & 0.786 & 0.097 & 0.765 & 0.103 & 0.760 & 0.127 & 0.801 & \tgb{0.070} & 0.765 & 0.076 & 0.765 & 0.116 \\
    & CAGVgg~\cite{mohammadi2020cagnet}  & 0.795 & 0.080 & \tbb{0.700} & \tbb{0.208} & 0.782 & 0.115 & 0.808 & 0.098 & 0.764 & 0.102 & 0.751 & 0.120 & 0.763 & 0.127 & 0.795 & 0.081 & 0.744 & 0.093 & 0.767 & 0.114\\
    & RASNet~\cite{chen2020reverse}  & \tgb{0.821} & \tbb{0.066} & 0.626 & 0.276 & 0.785 & \tbb{0.106} & \tgb{0.816} & \trb{0.087} & 0.788 & \tgb{0.086} & 0.776 & \trb{0.096} & 0.779 & \tgb{0.113} & \tbb{0.810} & \trb{0.066} & 0.774 & \trb{0.070} & 0.772 & 0.107 \\
    & CVAE~\cite{zhang2020uncertainty}  & 0.813 & 0.075 & 0.688 & 0.217 & \tgb{0.790} & \tgb{0.107} & \tgb{0.816} & \tgb{0.092} & 0.784 & 0.091 & 0.771 & 0.104 & 0.776 & 0.115 & \trb{0.820} & \tbb{0.069} & 0.767 & 0.080  & \tgb{0.781} & \tgb{0.106}\\
    & LDF~\cite{wei2020label}  & 0.819 & \tgb{0.071} & \tgb{0.697} & \tgb{0.212} & \tbb{0.796} & \trb{0.105} & \tbb{0.824} & \tbb{0.088} & \tgb{0.792} & \tbb{0.085} & \tbb{0.781} & \tbb{0.098} & \tbb{0.790} & \trb{0.107} & 0.780 & 0.073 & \trb{0.801} & \tbb{0.072} & \tbb{0.787} & \tbb{0.101} \\
    & R2Net~\cite{feng2020residual}  & \tbb{0.827} & \tgb{0.071} & 0.656 & 0.257 & \trb{0.802} & \tgb{0.107} & \trb{0.826} & \tgb{0.092} & \tbb{0.794} & 0.097 & \trb{0.789} & \tgb{0.099} & \trb{0.791} & \tbb{0.112} & \tgb{0.807} & 0.072 & \tbb{0.788} & \tgb{0.073} & \tbb{0.787} & 0.109 \\
    & BAS~\cite{qin2021boundary}  & \trb{0.831} & \trb{0.060} & \trb{0.723} & \trb{0.166} & 0.785 & 0.110 & 0.814 & 0.093 & \trb{0.797} & \trb{0.072} & \tgb{0.780} & 0.101 & \tgb{0.781} & 0.114 & \trb{0.820} & 0.072 & \tgb{0.787} & 0.075 & \trb{0.791} & \trb{0.096} \\
    \hline
    & Avg.  & 0.721 & 0.125 & 0.551 & 0.346 & 0.688 & 0.168 & 0.729 & 0.139 & 0.693 & 0.137 & 0.687 & 0.152 & 0.668 & 0.185 & 0.722 & 0.111 & 0.693 & 0.115 & - & - \\
   \bottomrule
  \end{tabular}
\end{table*}

In sharp contrast to the deep learning models,
the traditional models (\figref{fig:TraditionalTop50}) all fail
without exception.
WSC, HCCH, and RBD are the three most promising approaches.
However, their results are still far from the GT map,
since they are mainly based on various prior features extracted from color,
orientation, contrast, \etc.
Further, the center bias prior is not suitable in this case,
since the human is located close to the image boundary,
thus making this example more challenging for these approaches.

\section{Further Benchmarking}\label{sec:discussion}

\subsection{Attribute-Based Evaluation}
Based on the top-ranked models presented in Tables
\ref{tab:TraditionalSODScore} \& \ref{tab:DeepSODScore},
we further re-train the top-10\footnote{
DFI mode has only released the test code,
so we cannot evaluate it.}
deep learning models (using their default settings) on the
SOC-Sal\_train set (1,800 images) and then test them on the SOC-Sal\_test
set for attribute-based evaluation.
In \tabref{tab:AttributeBenchmark}, we show the performance
on subsets of our dataset characterized by a particular attribute.
Due to space limitations, in the following discussion, we only select a few
representative attributes for further analysis.

\textit{Big object} (BO) scenes typically occur when objects are close to
the camera, enabling tiny text and patterns to be seen clearly.
In this case, models that prefer to focus on local information
are seriously misled,
leading to a considerable decrease in performance
(\eg, 8.6\% $S_\alpha$ reduction for BAS, 8.7\% reduction for CAGVgg,
11.4\% reduction for LDF, and 40.7\% reduction for COV)
compared with their average performance (Avg.).
Among all attributes, BOs are the most difficult for both traditional
and deep learning models.

\textit{Small objects} (SOs) are tricky for some SOD models.
Four models (\ie, BAS, CVAE, CAGVgg, and RASNet) encounter performance
degradation (\eg, from BAS-0.5\% to RASNet-3.6\%) because SOs
are easily ignored during the downsampling of CNNs.
Other models instead have enhanced performance on SOs,
but significant reduction in performance on BOs.

\textit{Heterogeneous objects} (HOs) commonly appear in natural scenes.
The performance of all models on HOs improves to some degree, fluctuating from
2.9\% to 14.3\%.
We suspect this is because, as shown in \figref{fig:Attributes_distribution},
HO images make up a significant proportion of all datasets,
so the models are more familiar with this attribute.

\textit{Occlusion} (OC) scenes occur when objects are partly obscured.
Thus, SOD models must capture global semantics to make up for the
incomplete information of objects.
As observed, traditional models obtain improved performance compared with
their average performance.
For deep learning models, in contrast, this situation is reversed.

As can be seen in the last row of \tabref{tab:AttributeBenchmark}
(average performance of each attribute),
\textit{MB} and \textit{SO} have the same $S_\alpha$ score.
Moreover, the average scores of \textit{AC} and \textit{SC} are very similar.
It seems that existing deep learning based SOD models can effectively
address appearance change and shape complexity.
Similar to the attributes of \textit{OV} and \textit{OC}, \textit{CL} and
\textit{MB} remain challenging for existing methods,
generating mid-level (\ie, 0.65$<S_\alpha <$0.70) S-measure scores.

\begin{table}[t!]
  \centering
  \scriptsize
  \renewcommand{\arraystretch}{0.1}
  \renewcommand{\tabcolsep}{7.5mm}
  \caption{The contribution of our dataset-enhancement strategies.
  }\label{tab:effectiveness_of_our_dataset_strategies}
  \vspace{-10pt}
  \begin{tabular}{@{}r|ccccccc|cc|r}
  \toprule
  \diagbox[width=10em,height=2.0em,trim=l]{Method}{Metric} & $S_{\alpha}\uparrow$ & $E_{\xi}^{max}\uparrow$ & ${M}\downarrow$ \\
  \midrule
  ABP~\cite{zhang2020uncertainty} & 0.752 & 0.836 &  0.097\\
   {Our-ABP} & \textbf{0.769} &  \textbf{0.842} &  \textbf{0.093}\\
  \hline
   EGNet~\cite{zhao2019egnet} &  0.756 &  0.823 &  0.105\\
   Our-EGNet & \textbf{0.759} &  \textbf{0.831} &  \textbf{0.100}\\
  \hline
   CPDVgg~\cite{wu2019cascaded} &  0.775 &  0.842 &  0.090\\
   Our-CPDVgg &  \textbf{0.789} &  \textbf{0.850} &  \textbf{0.087}\\
  \hline
   CAGVgg~\cite{mohammadi2020cagnet}&  0.748 &  0.811 &  0.103\\
   Our-CAGVgg &  \textbf{0.759} &  \textbf{0.823} &  \textbf{0.097}\\
  \hline
  RASNet~\cite{chen2020reverse}  & 0.832 & 0.887 & 0.103 \\
  Our-RASNet & \textbf{0.841} & \textbf{0.897} & \textbf{0.096} \\
  \hline
  CVAE~\cite{zhang2020uncertainty}  & 0.849 & 0.892 & 0.089 \\
  Our-CVAE & \textbf{0.863} & \textbf{0.902} & \textbf{0.086} \\
  \hline
  LDF~\cite{wei2020label}  & 0.835 & 0.878 & 0.103 \\
  Our-LDF & \textbf{0.845} & \textbf{0.891} & \textbf{0.097} \\
  \hline
  R2Net~\cite{feng2020residual}  & 0.857 & 0.885 & 0.084 \\
  Our-R2Net & \textbf{0.868} & \textbf{0.899} & \textbf{0.080} \\
  \hline
  BAS~\cite{qin2021boundary}  & 0.842 & 0.882 & 0.092 \\
  Our-BAS & \textbf{0.856} & \textbf{0.895} & \textbf{0.086} \\
  \bottomrule
  \end{tabular}
\end{table}

\begin{table}[t!]
  \centering
  \scriptsize
  \renewcommand{\arraystretch}{0.1}
  \renewcommand{\tabcolsep}{7.5mm}
  \caption{The contribution of each dataset-enhancement strategy.
  } \label{tab:analysis_of_each_strategy}
  \vspace{-10pt}
  \begin{tabular}{@{}r|ccccccc|cc|r}
  \toprule
  \diagbox[width=10em,height=2.0em,trim=l]{Method}{Metric} & $S_{\alpha}\uparrow$ & $E_{\xi}^{max}\uparrow$ & ${M}\downarrow$ \\
  \midrule
  CVAE~\cite{zhang2020uncertainty}  & 0.849 & 0.892 & 0.089 \\
  \hline
  LS  & 0.851 & 0.895 & 0.088 \\
  SS  & 0.852 & 0.894 & 0.088 \\
  RDA & 0.855 & 0.896 & 0.086 \\
  \hline
  Our-CVAE & \textbf{0.863} & \textbf{0.902} & \textbf{0.086} \\
  \bottomrule
  \end{tabular}
\end{table}

\subsection{Comparison with Baselines}
We introduce three dataset-enhancement strategies to prevent networks from
being overconfident as a result of dataset bias.
These include label smoothing, random data augmentation and
self-supervised learning.
We argue that our strategies can be easily used in existing salient object
detection frameworks as general data processing techniques.
We thus introduce our strategies to nine benchmark salient object detection
models and show the performance in \tabref{tab:effectiveness_of_our_dataset_strategies},
where \enquote{Our-} represents the benchmark models with our dataset-enhancement strategies. Further, we investigate the contribution of each data-enhancement strategy, and show the performance in \tabref{tab:analysis_of_each_strategy}, where we choose CVAE~\cite{zhang2020uncertainty} as the base model.

\textbf{Training \& Testing Protocols.}
We retrain the five models in \tabref{tab:effectiveness_of_our_dataset_strategies} with their corresponding training dataset, \eg, MB~\cite{LiuSZTS07Learn} for RASNet~\cite{chen2020reverse}, and DUTS~\cite{wang2017learning} for all the other four models. We follow their original training and testing settings, \eg, same maximum epoch, learning rate, training and testing image sizes.

\textbf{Discussion.} \tabref{tab:effectiveness_of_our_dataset_strategies} shows consistent better performance of models with our strategies, which illustrates effectiveness of our solutions.
Further, in \tabref{tab:analysis_of_each_strategy},
\enquote{LS}, \enquote{RDA}, \enquote{SS} represent adding label smoothing
strategy, random data augmentation and self-supervised learning to the base model respectively. It shows that the random data augmentation achieves the largest performance gain, while label smoothing and self-supervised learning achieves comparable performance improvement. The main reason is that data augmentation introduce diverse samples to the initial training dataset, which is effective in improving model generalization ability.
For the self-supervised learning strategy, as the CVAE model \cite{zhang2020uncertainty} has already adopt the multi-scale image as input strategy, we observe slightly improved performance. However, the better performance in general can still validate the effectiveness of the proposed strategy.
Label smoothing \cite{rethink_inception} was introduced to prevent model from over-confidence, thus achieve well-calibrated model. However, there exists no saliency metrics to explain the calibration error of the saliency models. We will investigate in expected calibration error \cite{guo2017calibration} and extend it to saliency detection task in the future to better explain the calibration error issue. %of saliency models.

\begin{table}[t!]
  \centering
  \scriptsize
  \renewcommand{\arraystretch}{0.5}
  \renewcommand{\tabcolsep}{1.0mm}
  \caption{Results for cross-dataset generalization in \secref{sec:discussion}.
  UC-Net (CVPR'20)~\cite{zhang2020uc} is trained on one dataset and tested on all others. ``Sel.'': diagonal score (training and testing on the same dataset).
  ``Oth.'': mean score on all except for self.
  }\label{tab:crossDataset_UCNet}
  \vspace{-10pt}
  \begin{tabular}{@{}r|ccccccc|cc|r}
  \toprule
  Measure & \multicolumn{9}{c|}{$S_{\alpha}\uparrow$~\cite{fan2017structure}} & Drop$\downarrow$\\
  \midrule
  \diagbox[width=5em,height=1.6em,trim=l]{Train}{Test} & SOC & M10K  & DU-O & DUTS & ECC & HKU & ILSO &Sel. & Oth. & \\
    \midrule
     SOC~\cite{fan2018salient}  & \textbf{.884} & .768 & .686 & .834  & .749 & .774
     & .841 & .884 & .775 &  12\%\\
     M10K~\cite{ChengPAMI15}   & .800 & \textbf{.921}& .784 & .894& .881& .882& .884
     & .921 & .854 & 7\% \\
     DU-O~\cite{YangZLRY13Manifold} &.833 &.898  &\textbf{.854}  & .877 & .862&.867 & .886 & .854 & .871 & -2\%\\
     DUTS~\cite{wang2017learning} &.795 & .882 & .793&\textbf{.910} & .890&.903 & .900
      & .910 &.861 & 5\% \\
     ECC~\cite{yan2013hierarchical}  &.791 & .886 &.800 &.901 &\textbf{.901} & .898& .903  & .901 & .863 & 4\%\\
     HKU~\cite{li2015visual} & .818 &.892 & .787&.904 & .883& \textbf{.910} & .905
      & .910 & .865 & 5\%\\
     ILSO~\cite{li2017instance}  &.841 & .888 &.790 & .898& .882&.896 & \textbf{.920}& .920 & .866 & 6\%\\
     \midrule
     Oth. & .813 & .869 & .773 & .885 & .858 & .870 & .887 & &\\
  \bottomrule
  \end{tabular}
\end{table}

\subsection{Cross-Dataset Generalization}
To study the difficulty of existing SOD datasets,
we adopt the CDA (cross-data analysis) method~\cite{torralba2011unbiased}.
Given $N$ candidate datasets $\{D^n\}_{n=1}^{N}$,
we first train a model on the $D_i$  dataset,
and then test it on the other datasets (\ie, $\{D^n\}_{n=1,n \neq i}^{N}$).
Following~\cite{fan2021concealed,wang2021salient},
we randomly select 800 images and 200 images from each dataset
as the training set and testing set, respectively.

We train the representative UC-Net~\cite{zhang2020uc} (CVPR2020 Best Paper Nomination) on
existing popular datasets that contain more than 1,000 images.
\tabref{tab:crossDataset_UCNet} shows the $S_\alpha$ score on each dataset.
Each column provides the score of UC-Net tested on a specific dataset and
trained on all others.
Each row indicates the performance of UC-Net trained on one dataset and
tested on all others,
demonstrating the generalizability of the dataset adopted for training.
We find that when testing on our SOC (\eg, Oth. = 0.813) and DU-O (Oth. = 0.773)
datasets, the model performs worse than other datasets.
It shows larger differences between SOC/DU-O and the other datasets.

\section{Future Directions}\label{sec:futureworks}

Human attention can be influenced by four key factors:
\begin{itemize}
  \item \textbf{Visual properties}.
    Our attention may be drawn by basic objects' unique visual
    properties~\cite{cave2017finding}.
  \item \textbf{Memory}.
    If one knows an object well,
    it is easier for that object to attract one's attention.
  \item \textbf{Goal}.
    For example, eye fixation records are quite different from attention maps,
    with a specific goal for viewers.
  \item \textbf{Emotion.}
    In addition to the above-mentioned factors,
    we argue that human attention toward the same scene may be affected
    by one's emotion, \eg, happiness, sadness, anger.
\end{itemize}

As demonstrated by Cave~\cite{cave2017finding},
attentional control is determined by a combination of these factors.
Unfortunately, the annotations of existing SOD datasets do not clearly describe which factor they address.
Differently, the ground-truth annotations of our SOC are based on
the salicon (free-view task) dataset\footnote{\url{http://salicon.net/}},
or so-call meaning maps which are used in recent
studies~\cite{cave2017finding,henderson2019meaning,henderson2017meaning}.
As concluded by Kalash \etal~\cite{kalash2019relative},
the work to date has addressed a relatively ill-posed problem.
Thus, we recommend several future directions to re-think SOD tasks
at six main research levels:

\textbf{(1) Data Level:}
Recently, visual saliency detection tasks have attracted
significant interest using 2D (RGB SOD) and 3D (\ie, RGB-D, RGB-T)
input data.
However, light field SOD (4D), LIDAR SOD, and 360$^\circ$ SOD are still not well-studied.
Establishing new datasets for these types of data will largely promote the
development of this field.
Another interesting avenue for examining saliency detection is to
study fine-grained tasks,
such as salient instance detection
\cite{li2017instance,fan2019s4net,tian2020weakly,wu2021regularized} and part-object visual saliency detection~\cite{liu2021part}.

\textbf{(2) Task Level:}
Multi-task learning has demonstrated strong performance in recent works
\cite{zamir2018taskonomy}.
Existing schemes mainly focus on vision tasks,
such as joint salient object detection and camouflaged object detection
\cite{aixuan_cod_sod21},
detection of salient objects, edges and skeletons simultaneously
\cite{liu2020dynamic},
and simultaneous detection, ranking, and subitizing of multiple salient
objects~\cite{islam2018revisiting}.
With the success of the transformer technique in natural language processing
(NLP), introducing multi-modality learning into the saliency detection field
may be a feasible way to further incorporate other types of information,
such as CV+NLP (similar to~\cite{Zhuge2021KaleidoBERT}),
CV+Audio~\cite{from2021wang}, and CV+other modality.

\textbf{(2) Model Level:}
A huge number of algorithms have been developed to improve detection accuracy.
However, there are several promising directions that could be further studied
such as data augmentation techniques~\cite{ruiz2020ida},
efficient SOD models
(\eg, lightweight models~\cite{gao2020highly,liu2021samnet}),
new loss functions~\cite{chen2020contour,21SC-Emeasure},
ranking-based models~\cite{zhang2017ranking,islam2018revisiting},
and transformer-based models~\cite{mao2021transformer,liu2021VST}.

\textbf{(4) Supervision Level:}
In addition to the most common fully supervised learning of current SOD models,
other supervision strategies, \eg, weakly supervised
(\ie, scribble~\cite{zhang2020weakly}, category~\cite{zhang2021learning},
and polygon),
semi-supervised~\cite{zhou2019semisu},
self-supervised~\cite{wang2019retrieval,zhao2021self},
and unsupervised~\cite{zhang2018deep} learning are also interesting to study.

\textbf{(5) Evaluation Level:}
Evaluation metrics are important for model training, testing, and
benchmarking.
However, the SOD community still utilizes classical metrics such as IoU,
F-measure, and MAE.
These metrics were designed for universal evaluation rather than for assessing SOD tasks
specifically.
As a consequence, they do not work well for certain specific applications,
such as those with high-quality requirements.
We envision that introducing a new metric
(\eg, based on the gradient or connectivity error used in~\cite{deora2021salient})
for SOD tasks,
such as weighted F-measure~\cite{margolin2014evaluate} and
S-measure~\cite{fan2017structure}, will be another important research direction.

\textbf{(6) Application Level:}
The SOD task belongs to a more general task called class-agnostic
object detection (CAOD)~\cite{jia2013category}.
For simple scenes (\eg, those containing only one or two clear objects),
SOD is identical to CAOD.
From this point of view, SOD models have many potential applications
in the real-world
(\eg, Alibaba's fashion search system~\cite{Zhuge2021KaleidoBERT}),
despite their currently limited number of representative cases
\cite{kumar2012leafsnap,qin2021boundary,hou2017dss}.

\section{Conclusion}
In this survey, we identified and addressed the long-ignored
\textit{data selection bias} issue in SOD.
Different from previous studies, we aimed to explore the SOD task in the wild.
To achieve this goal, we collected a new challenging and densely annotated
\ourdataset~dataset;
analyzed a large number ($\sim$200) of representative models;
conducted the most complete (\ie, 100) benchmarking;
devised a series of simple learning strategies to efficiently utilize negative
samples and training data;
and identified several current challenges and future directions.
We hope that these contributions will provide the SOD community an
opportunity to explore novel techniques in an open environment.
We have tried to cover the most important works.
Nevertheless, it is impractical to thoroughly investigate all models in
this vast field.
We will continue to incorporate new techniques on our website.

% \myPara{Acknowledgments.}
% %The authors would like to thank the anonymous reviewers and editor for their helpful comments on this manuscript.
% This work is funded by the National Key Research and Development Program
% of China (No.2018AAA0100400)
% and NSFC (NO. 61922046).

{
\bibliographystyle{IEEEtran}
\bibliography{SOC}
}

\end{document}